\documentclass[10pt,twocolumn,letterpaper]{article}

\usepackage{cvpr}
\usepackage{times}
\usepackage{epsfig}
\usepackage{graphicx}
\usepackage{amsmath}
\usepackage{amssymb}
\usepackage{multirow}
\usepackage{subfig}
\usepackage[title]{appendix}
\usepackage{xspace}
\usepackage{color}
\usepackage{multirow}
\usepackage{adjustbox}
\usepackage{paralist} 
\usepackage{enumitem} 
\usepackage{booktabs} 
\usepackage{array}
\usepackage{caption} 

\graphicspath{{figures/}}

\newcommand{\Paragraph}[1]{{\flushleft{\textbf{#1}}}} 

\newcommand{\tb}[1]{\textbf{#1}}

\definecolor{gray}{rgb}{0.35,0.35,0.35}
\definecolor{MyBlue}{rgb}{0,0.2,0.8}
\definecolor{MyRed}{rgb}{0.8,0.2,0}
\definecolor{MyGreen}{rgb}{0.0,0.4,0.1}
\definecolor{MyGray}{rgb}{0.4,0.4,0.4}

\long\def\ignorethis#1{}

\newlength\paramargin
\newlength\figmargin
\newlength\secmargin

\newcolumntype{L}[1]{>{\raggedright\let\newline\\\arraybackslash\hspace{0pt}}m{#1}}
\newcolumntype{C}[1]{>{\centering\let\newline\\\arraybackslash\hspace{0pt}}m{#1}}
\newcolumntype{R}[1]{>{\raggedleft\let\newline\\\arraybackslash\hspace{0pt}}m{#1}}

\def\eg{\textit{e.g.},\xspace}

\def\etal{et~al.\xspace}

\newcommand{\figref}[1]{Figure~\ref{fig:#1}}

\usepackage[pagebackref=true,breaklinks=true,letterpaper=true,colorlinks,bookmarks=false,citecolor=blue,linkcolor=blue, urlcolor=blue]{hyperref}

\cvprfinalcopy 

\newcommand\blfootnote[1]{%
  \begingroup
  \renewcommand\thefootnote{}\footnote{#1}%
  \addtocounter{footnote}{-1}%
  \endgroup
}

\def\cvprPaperID{869} 

\makeatletter
\newcommand{\printfnsymbol}[1]{%
  \textsuperscript{\@fnsymbol{#1}}%
}
\makeatother
\begin{document}

\title{Mode Seeking Generative Adversarial Networks for Diverse Image Synthesis}

\author{Qi Mao\printfnsymbol{1}$^1$, Hsin-Ying Lee\printfnsymbol{1}$^2$, Hung-Yu Tseng\printfnsymbol{1}$^2$, Siwei Ma$^{1,3}$, Ming-Hsuan Yang$^{2,4}$ \vspace{2mm}\\
$^{1}$Institute of Digital Media, Peking University\hspace{20pt}$^{2}$University of California, Merced\\
$^{3}$Peng Cheng Laboratory\hspace{20pt}$^{4}$Google Cloud}

\twocolumn[{%
\renewcommand\twocolumn[1][]{#1}%
\maketitle
\begin{center}
    \vspace{-4.2mm}
	\includegraphics[width=0.99\textwidth]{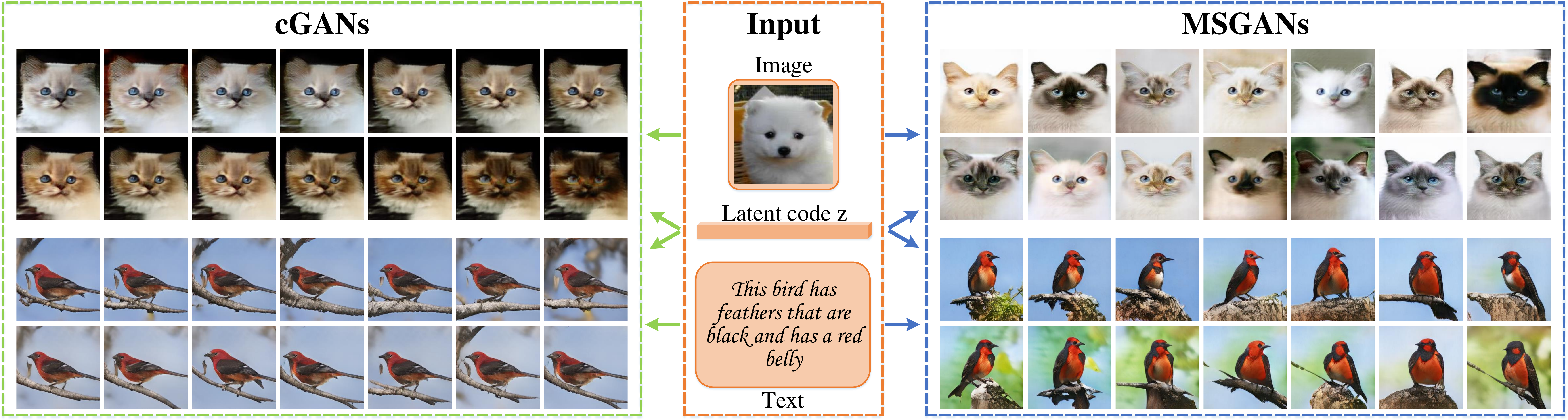}
	\vspace{-3mm}
	\captionof{figure}{\textbf{Mode seeking generative adversarial networks~(MSGANs).} 
	(\textit{Left}) Existing conditional generative adversarial networks tend to ignore the input latent code $\mathbf{z}$ and generate images of similar modes. 
	(\textit{Right}) We propose a simple yet effective mode seeking regularization term that can be applied to arbitrary conditional generative adversarial networks in different tasks to alleviate the mode collapse issue and improve the diversity.}
	\label{fig:all_results}
\end{center}
}]

\begin{abstract}
\vspace{-2mm}
\blfootnote{$\ast$ Equal contribution}
Most conditional generation tasks expect diverse outputs given a single conditional context.
However, conditional generative adversarial networks (cGANs) often focus on the prior conditional information and ignore the input noise vectors, which contribute to the output variations.
Recent attempts to resolve the mode collapse issue for cGANs are usually task-specific and computationally expensive.
In this work, we propose a simple yet effective regularization term to address the mode collapse issue for cGANs.
The proposed method explicitly maximizes the ratio of the distance between generated images with respect to the corresponding latent codes, thus encouraging the generators to explore more minor modes during training.
This mode seeking regularization term is readily applicable to various conditional generation tasks without imposing training overhead or modifying the original network structures.
We validate the proposed algorithm on three conditional image synthesis tasks including categorical generation, image-to-image translation, and text-to-image synthesis with different baseline models.
Both qualitative and quantitative results demonstrate the effectiveness of the proposed regularization method for improving diversity without loss of quality.
\end{abstract}

\section{Introduction}
Generative adversarial networks (GANs)~\cite{goodfellow2014generative} have been shown to capture complex and high-dimensional image data with numerous applications effectively. 
Built upon GANs, conditional GANs (cGANs)~\cite{mirza2014conditional} take external information as additional inputs.
For image synthesis, cGANs can be applied to various tasks with different conditional contexts.
With class labels, cGANs can be applied to categorical image generation.  
With text sentences, cGANs can be applied to text-to-image synthesis~\cite{reed2016generative, zhang2017stackgan++}. 
With images, cGANs have been used in tasks including image-to-image translation~\cite{huang2018multimodal,isola2017image, lee2018diverse, liu2017unsupervised,zhu2017unpaired, zhu2017toward}, semantic manipulation~\cite{wang2018pix2pixHD} and style transfer~\cite{li2016precomputed}.

For most conditional generation tasks, the mappings are in nature multimodal, i.e., a single input context corresponds to multiple plausible outputs.
A straightforward approach to handle multimodality is to take random noise vectors along with the conditional contexts as inputs, where the contexts determine the main content and noise vectors are responsible for variations.
For instance, in the dog-to-cat image-to-image translation task~\cite{lee2018diverse}, the input dog images decide contents like orientations of heads and positions of facial landmarks, while the noise vectors help the generation of different species.
However, cGANs usually suffer from the \textit{mode collapse}~\cite{goodfellow2014generative,salimans2016improved} problem, where generators only produce samples from a single or few modes of the distribution and ignore other modes.
The noise vectors are ignored or of minor impacts, since cGANs pay more attention to learn from the high-dimensional and structured 
conditional contexts.

There are two main approaches to address the mode collapse problem in GANs.
A number of methods focus on discriminators by introducing different divergence metrics~\cite{arjovsky2017wasserstein,mao2017least} and optimization process~\cite{durugkar2017generative,metz2016unrolled,salimans2016improved}.
The other methods use auxiliary networks such as multiple generators~\cite{ghosh2018multi, liu2016coupled} and additional encoders ~\cite{che2016mode, donahue2016adversarial, dumoulin2016adversarially, srivastava2017veegan}.
However, mode collapse is relatively less studied in cGANs.
Some recent efforts have been made in the image-to-image translation task to improve diversity~\cite{huang2018multimodal,lee2018diverse,zhu2017toward}. 
Similar to the second category with the unconditional setting, these approaches introduce additional encoders and loss functions to encourage the one-to-one relationship between the output and the latent code.
These methods either entail heavy computational overheads on training or require auxiliary networks that are often task-specific that cannot be easily extended to other frameworks.

In this work, we propose a mode seeking regularization method that can be applied to cGANs for various tasks to alleviate the mode collapse problem. 
Given two latent vectors and the corresponding output images, we propose to maximize the ratio of the distance between images with respect to the distance between latent vectors.
In other words, this regularization term encourages generators to generate dissimilar images during training.
As a result, generators can explore the target distribution, and enhance the chances of generating samples from different modes.
On the other hand, we can train the discriminators with dissimilar generated samples to provide gradients from minor modes that are likely to be ignored otherwise. 
This mode seeking regularization method incurs marginal computational overheads and can be easily embedded in different cGAN frameworks to improve the diversity of synthesized images. 

We validate the proposed regularization algorithm through an extensive evaluation of three conditional image synthesis tasks with different baseline models.
First, for categorical image generation, we apply the proposed method on DCGAN~\cite{radford2015unsupervised} using the CIFAR-10~\cite{krizhevsky2009learning} dataset.
Second, for image-to-image translation, we embed the proposed regularization scheme in Pix2Pix~\cite{isola2017image} and DRIT~\cite{lee2018diverse} using the facades~\cite{cordts2016cityscapes}, maps~\cite{isola2017image}, Yosemite~\cite{zhu2017unpaired}, and cat$\rightleftharpoons$dog~\cite{lee2018diverse} datasets.
Third, for text-to-image synthesis, we incorporate StackGAN++~\cite{zhang2017stackgan++} with the proposed regularization term using the CUB-200-2011~\cite{WahCUB_200_2011} dataset.
We evaluate the diversity of synthesized images using perceptual distance metrics~\cite{zhang2018unreasonable}.

However, the diversity metric alone cannot guarantee the similarity between the distribution of generated images and the distribution of real data.
Therefore, we adopt two recently proposed bin-based metrics~\cite{richardson2018NDB}, the \textit{Number of Statistically-Different Bins} (NDB) metric which determines the relative proportions of samples fallen into clusters predetermined by real data, and the \textit{Jensen-Shannon Divergence} (JSD) distance which measures the similarity between bin distributions.
Furthermore, to verify that we do not achieve diversity at the expense of realism, we evaluate our method with the Fr\'{e}chet Inception Distance (FID)~\cite{heusel2017gans}
as the metric for quality. 
Experimental results demonstrate that the proposed regularization method can facilitate existing models from various applications achieving better diversity without loss of image quality. 
\figref{all_results} shows the effectiveness of the proposed regularization method for existing models.

The main contributions of this work are:
\begin{compactitem}
\item 
We propose a simple yet effective mode seeking regularization method to address the mode collapse problem in cGANs.
This regularization scheme can be readily extended into existing frameworks with marginal training overheads and modifications.
\item We demonstrate the generalizability of the proposed regularization method on three different conditional generation tasks: categorical generation, image-to-image translation, and text-to-image synthesis.
\item 
Extensive experiments show that the proposed method can facilitate existing models from different tasks achieving better diversity without sacrificing visual quality of the generated images. 
\end{compactitem}

Our code and pre-trained models are available at \url{https://github.com/HelenMao/MSGAN/}. 
\begin{figure*}[t]
\begin{center}
	\includegraphics[width=1\linewidth]{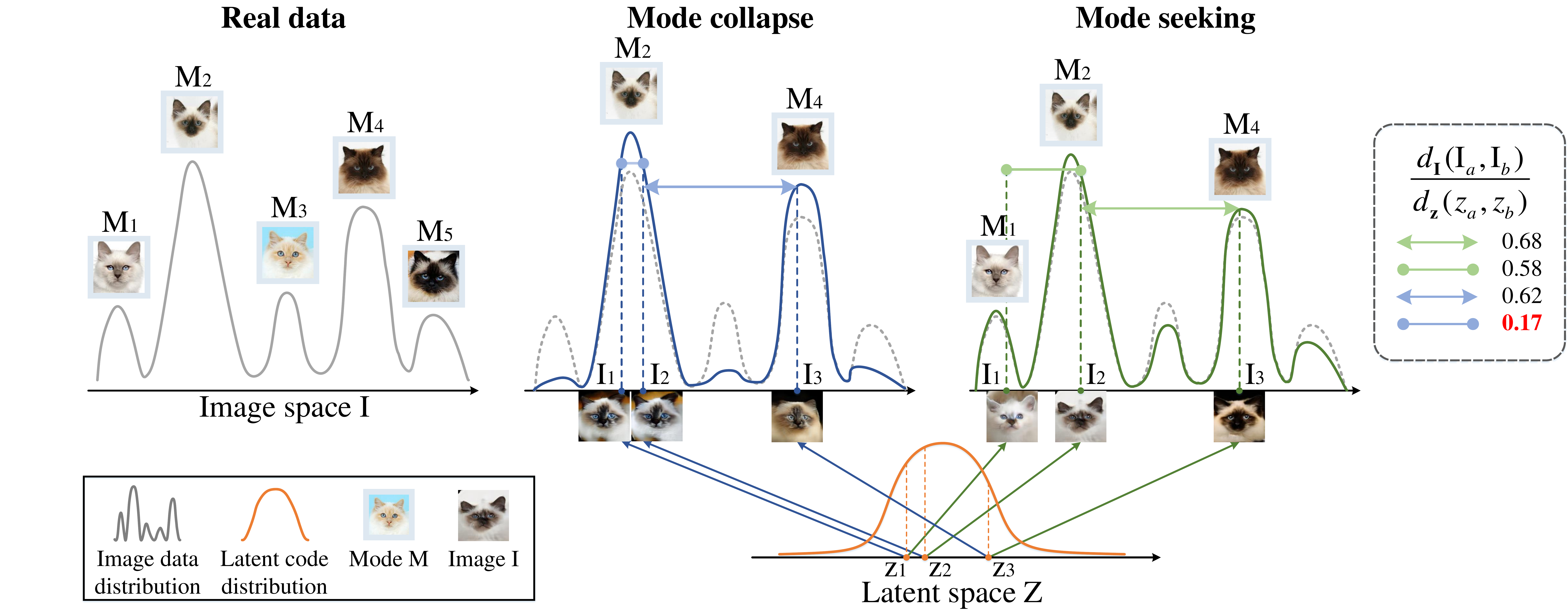}
	\vspace{-1mm}
\caption{
	\textbf{Illustration of motivation.}
	Real data distribution contains numerous modes.
	However, when mode collapse occurs, generators only produce samples from a few modes.
	From the data distribution when mode collapse occurs, we observe that for latent vectors $\mathbf{z_1}$ and $\mathbf{z_2}$, the distance between their mapped images $\mathbf{I_1}$ and $\mathbf{I_2}$ will become shorter in a disproportionate rate when the distance between two latent vectors is decreasing.
	We present on the right the ratio of the distance between images with respect to the distance of the corresponding latent vectors, where we can spot an anomalous case (colored in red) where mode collapse occurs.
	The observation motivates us to leverage the ratio as the training objective explicitly.
	}
	\label{fig:motivation}
	\vspace{-5mm}
\end{center}
\end{figure*}

\section{Related Work}
\vspace{-2mm}
\paragraph{Conditional generative adversarial networks.}
Generative adversarial networks~\cite{arjovsky2017wasserstein, goodfellow2014generative,mao2017least,radford2015unsupervised} have been widely used for image synthesis. 
With adversarial training, generators are encouraged to capture the distribution of real images.
On the basis of GANs, \textit{conditional} GANs synthesize images based on various contexts.
For instances, cGANs can generate high-resolution images conditioned on low-resolution images~\cite{ledig2017photo}, translate images between different visual domains~\cite{huang2018multimodal,isola2017image, lee2018diverse, liu2017unsupervised,zhu2017unpaired, zhu2017toward}, generate images with desired style~\cite{li2016precomputed}, and synthesize images according to sentences~\cite{reed2016generative,zhang2017stackgan++}.
Although cGANs have achieved success in various applications, existing approaches suffer from the mode collapse problem.
Since the conditional contexts provide strong structural prior information for the output images and have higher dimensions than the input noise vectors, generators tend to ignore the input noise vectors, which are responsible for the variation of generated images.
As a result, the generators are prone to produce images with similar appearances.
In this work, we aim to address the mode collapse problem for cGANs.
\vspace{-4mm}
\paragraph{Reducing mode collapse.}
Some methods focus on the discriminator with
different optimization process~\cite{metz2016unrolled} and divergence metrics~\cite{arjovsky2017wasserstein,mao2017least} to stabilize the training process. 
The minibatch discrimination scheme~\cite{salimans2016improved} allows the discriminator to discriminate between whole mini-batches of samples instead of between individual samples.
In~\cite{durugkar2017generative}, Durugkar~\etal~use multiple discriminators to address this issue. 
The other methods use auxiliary networks to alleviate the mode collapse issue. 
ModeGAN~\cite{che2016mode} and VEEGAN~\cite{srivastava2017veegan} enforce the bijection mapping between the input noise vectors and generated images with additional encoder networks.
Multiple generators~\cite{ghosh2018multi} and weight-sharing generators~\cite{liu2016coupled} are developed to capture more modes of the distribution.
However, these approaches either entail heavy computational overheads or require modifications of the network structure, and may not be easily applicable to cGANs.
\vspace{-1mm}

In the field of cGANs, some efforts~\cite{huang2018multimodal,lee2018diverse,zhu2017toward} have been recently made to address the mode collapse issue on the image-to-image translation task.
Similar to ModeGAN and VEEGAN,
additional encoders are introduced to provide a bijection constraint between the generated images and input noise vectors.
However, these approaches require other task-specific networks and objective functions.
The additional components make the methods less generalizable and incur extra computational loads on training.
In contrast, we propose a simple regularization term that imposes no training overheads and requires no modifications of the network structure.
Therefore, the proposed method can be readily applied to various conditional generation tasks.
Recently, the concurrent work~\cite{yang2019diversity} also adopts a loss term similar to our work for reducing mode collapse for cGANs.
\vspace{-1mm}
\section{Diverse Conditional Image Synthesis}

\subsection{Preliminaries}
\vspace{-1mm}
The training process of GANs can be formulated as a mini-max problem: a discriminator $D$ learns to be a classifier by assigning higher discriminative values to the real data samples and lower ones to the generated ones.
Meanwhile, a generator $G$ aims to fool $D$ by synthesizing realistic examples.
Through adversarial training, the gradients from $D$ will guide $G$ toward generating samples with the distribution similar to the real data one.

The mode collapse problem with GANs is well known in the literature. 
Several methods~\cite{che2016mode,salimans2016improved, srivastava2017veegan} attribute the missing mode to the lack of penalty when this issue occurs.
Since all modes usually have similar discriminative values, larger modes are likely to be favored through the training process based on gradient descent. 
On the other hand, it is difficult to generate samples from minor modes.

The mode missing problem becomes worse in cGANs.
Generally, conditional contexts are high-dimensional and structured (\eg images and sentences) as opposed to the noise vectors.
As such, the generators are likely to focus on the contexts and ignore the noise vectors, which account for diversity.

\subsection{Mode Seeking GANs}
In this work, we propose to alleviate the missing mode problem from the generator perspective.
\figref{motivation} illustrates the main ideas of our approach.
Let a latent vector $\mathbf{z}$ from the latent code space $Z$ be mapped to the image space $I$.
When mode collapse occurs, the mapped images are collapsed into a few modes.
Furthermore, when two latent codes $\mathbf{z}_1$ and $\mathbf{z}_2$ are closer, the mapped images $\mathbf{I}_1=G(c, \mathbf{z}_1)$ and $\mathbf{I}_2=G(c, \mathbf{z}_2)$ are more likely to be collapsed into the same mode.
To address this issue, we propose a mode seeking regularization term to directly maximize the ratio of the distance between $G(c, \mathbf{z}_1)$ and $G(c, \mathbf{z}_2)$ with respect to the distance between $\mathbf{z}_1$ and $\mathbf{z}_2$, 
\begin{equation}
    \mathcal{L}_{\mathrm{ms}}= \max\limits_{G} (\frac{d_\mathbf{I}(G(c, \mathbf{z}_1), G(c, \mathbf{z}_2))}{d_\mathbf{z}(\mathbf{z}_1,\mathbf{z}_2)}),
\label{Eq:proposed1}  
\end{equation}
where $d_{\ast}(\cdot)$ denotes the distance metric.

The regularization term offers a virtuous circle for training cGANs.
It encourages the generator to explore the image space and enhances the chances for generating samples of minor modes.
On the other hand, the discriminator is forced to pay attention
to generated samples from minor modes. 
\figref{motivation} shows a mode collapse situation where two close samples, $\mathbf{z}_1$ and $\mathbf{z}_2$, are 
mapped onto the same mode $M_2$.
However, with the proposed regularization term, $\mathbf{z}_1$ is mapped to $\mathbf{I}_1$, which belongs to an unexplored mode $M_1$.
With the adversarial mechanism, the generator will thus have better chances to generate samples of $M_1$ in the following training steps.

\begin{figure}[t]
\begin{center}
\subfloat[Proposed regularization]{
\includegraphics[width=1\linewidth]{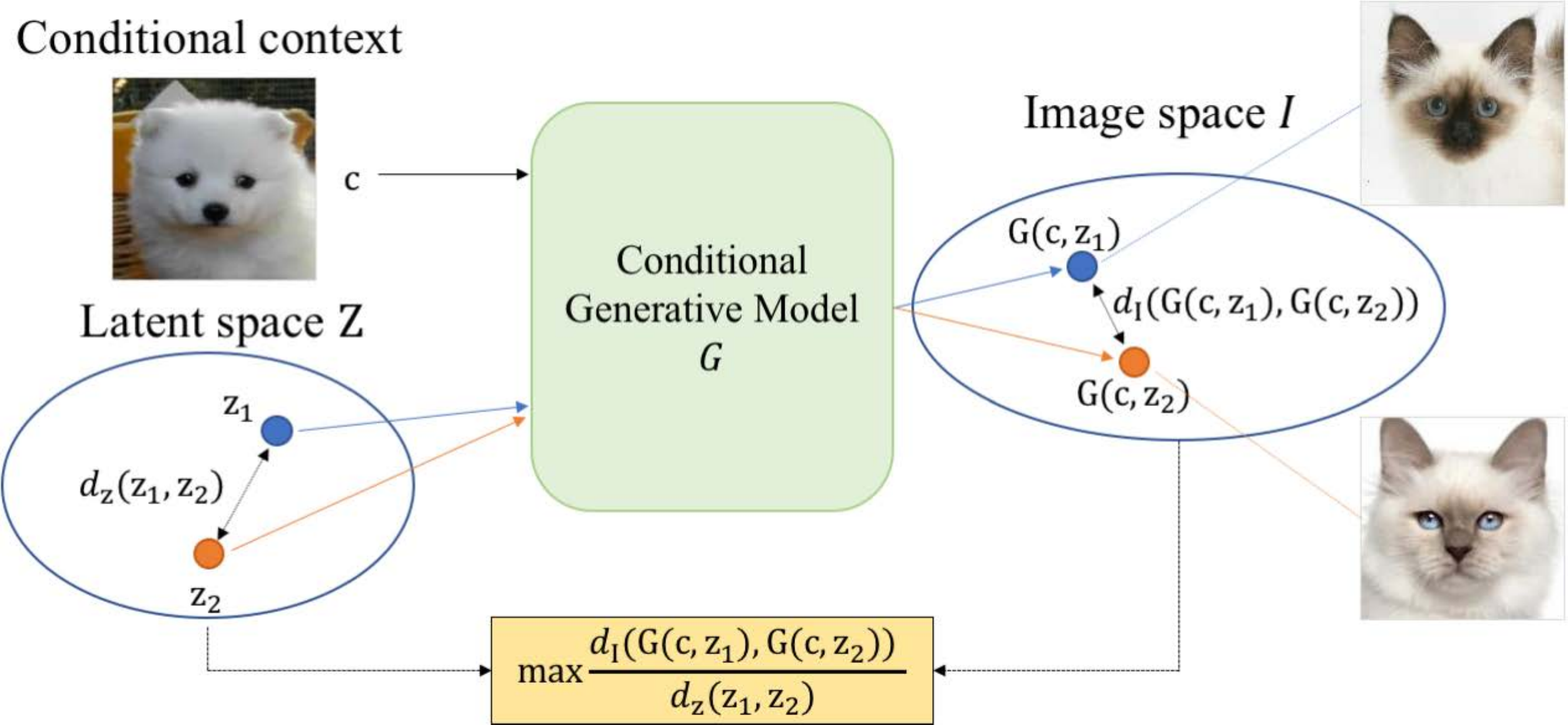}
}

\subfloat[Applying proposed regularization on StackGAN++]{
\includegraphics[width=1\linewidth]{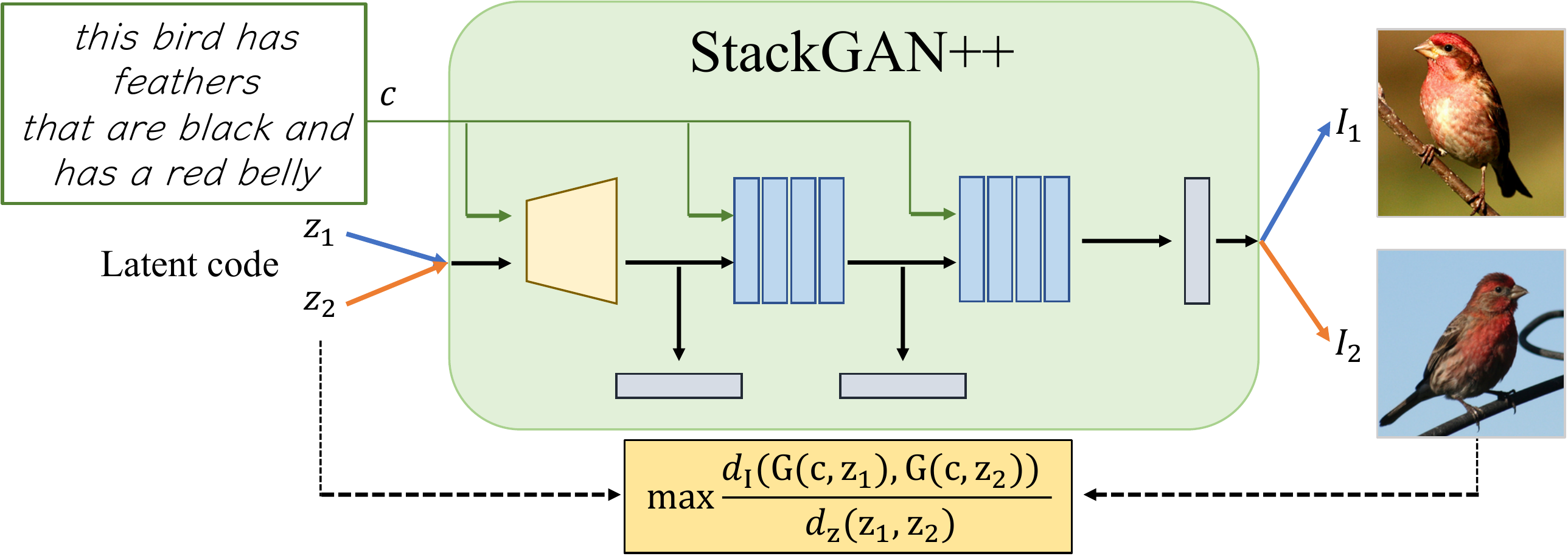}
}

\end{center}
\caption{\textbf{Proposed regularization.} 
(a) We propose a regularization term that maximizes the ratio of the distance between generated images with respect to the distance between their corresponding input latent codes.
(b) The proposed regularization method can be applied to arbitrary cGANs.
Take StackGAN++~\cite{zhang2017stackgan++}, a model for text-to-image synthesis, as an example, we easily apply the regularization term regardless of the complex tree-like structure of the original model.}
\label{fig:framework}
\vspace{-4mm}
\end{figure}

As shown in \figref{framework}, the proposed regularization term can be easily integrated with existing cGANs by appending it to the original objective function.
\begin{equation}
\mathcal{L}_{\mathrm{new}} = \mathcal{L}_{\mathrm{ori}} +\lambda_{\mathrm{ms}}\mathcal{L}_{\mathrm{ms}},
\end{equation}
where $\mathcal{L}_{\mathrm{ori}}$ denotes the original objective function and $\lambda_{\mathrm{ms}}$ the weights to control the importance of the regularization. 
Here, $\mathcal{L}_{\mathrm{ori}}$ can be as a simple loss function.
For example, in categorical generation task,
\begin{equation}
    \mathcal{L}_{\mathrm{ori}} = \mathbb{E}_{c,\mathbf{y}}[\log{D(c,\mathbf{y})}] + \mathbb{E}_{c,\mathbf{z}}[\log{(1-D(c,G(c,\mathbf{z})))}],
    \label{Eq:class}
\end{equation}
where $c,\mathbf{y},\mathbf{z}$ denote class labels, real images, and noise vectors, respectively.
In image-to-image translation task~\cite{isola2017image}, 
\begin{equation}
    \mathcal{L}_{\mathrm{ori}} = \mathcal{L}_{\mathrm{GAN}} + \mathbb{E}_{\mathbf{x},\mathbf{y},\mathbf{z}}[\left\lVert{\mathbf{y}-G(\mathbf{x},\mathbf{z})} \right\rVert_{1}],
    \label{Eq:I2I}
\end{equation}
where $\mathbf{x}$ denotes input images and $\mathcal{L}_{\mathrm{GAN}}$ is the typical GAN loss.
$\mathcal{L}_{\mathrm{ori}}$ can be arbitrary complex objective function from any task, as shown in \figref{framework} (b).
We name the proposed method as Mode Seeking GANs (MSGANs).

\section{Experiments}

We evaluate the proposed regularization method through extensive quantitative and qualitative evaluation.
We apply MSGANs to the baseline models from three representative conditional image synthesis tasks: categorical generation, image-to-image translation, and text-to-image synthesis.
Note that we augment the original objective functions with the proposed regularization term while maintaining original network architectures and hyper-parameters.
We employ $L_1$ norm distance as our distance metrics for both $d_\mathbf{I}$ and $d_\mathbf{z}$ and set the hyper-parameter $\lambda_{\mathrm{ms}}=1$ in all experiments.
More implementation and evaluation details, please refer to the appendixes.

\begin{table*}[t]
\centering
\caption{\textbf{NDB and JSD results on the CIFAR-10 dataset.} 
}
\vspace{-1mm}
\begin{tabular}{@{}ccccccc@{}} 
	    \toprule 
		Metrics & Models & airplane & automobile & bird & cat & deer \\ \cline{3-7}
        \multirow{2}{*}{NDB~$\downarrow$} &DCGAN & $49.60\pm{3.43}$ & $53.00\pm{7.28}$ &$\mathbf{34.40\pm{6.11}}$ & $46.00\pm{1.41}$& $\mathbf{44.80\pm{3.90}}$\\
        \multirow{2}{*}{ } &MSGAN &$\mathbf{46.60\pm{7.40}}$  &  $\mathbf{51.80\pm{2.28}}$ & $39.40\pm{1.95}$&$\mathbf{41.80\pm{3.70}}$&$46.80\pm{4.92}$ \\

        \midrule
         \multirow{2}{*}{JS~$\downarrow$} &DCGAN & $0.034\pm{0.001}$ &$0.035\pm{0.002}$ &$\mathbf{0.025\pm{0.002}}$ & $0.030\pm{0.002}$ &
         $\mathbf{0.033\pm{0.001}}$\\
        \multirow{2}{*}{ } &MSGAN &$\mathbf{0.031\pm{0.001}}$  &$\mathbf{0.033\pm{0.001}}$ & $0.027\pm{0.001}$ & $\mathbf{0.027\pm{0.001}}$&$0.035\pm{0.003}$\\
        \midrule
        		   &    & dog & frog & horse & ship & truck \\ \cline{3-7}
        \multirow{2}{*}{NDB~$\downarrow$} &DCGAN & $50.40\pm{4.62}$ & $52.00\pm{3.81}$ &$54.40\pm{4.04}$ & $42.80\pm{5.45}$&$47.80\pm{4.55}$\\
        \multirow{2}{*}{ } &MSGAN &$\mathbf{33.80\pm{3.27}}$  & $\mathbf{42.00\pm{2.92}}$ & $\mathbf{47.60\pm{5.03}}$&$\mathbf{41.00\pm{2.92}}$&$\mathbf{43.80\pm{6.61}}$ \\
        \midrule
         \multirow{2}{*}{JS~$\downarrow$} &DCGAN & $0.033\pm{0.001}$ &$0.034\pm{0.002}$ &$0.035\pm{0.001}$ & $0.029\pm{0.003}$ &${0.032}\pm{0.001}$ \\
        \multirow{2}{*}{ } &MSGAN &$\mathbf{0.024\pm{0.001}}$  &$\mathbf{0.030\pm{0.002}}$ & $\mathbf{0.033\pm{0.003}}$ & $\mathbf{0.027\pm{0.001}}$&$\mathbf{0.029\pm{0.003}}$\\
		\bottomrule
		
\end{tabular}
\label{tab:NDB_cifar}
\end{table*}

\subsection{Evaluation Metrics}
We conduct evaluations using the following metrics.
\noindent
\tb{FID. }
To evaluate the quality of the generated images, we use FID~\cite{heusel2017gans} to measure the distance between the generated distribution and the real one through features extracted by Inception Network~\cite{szegedy2015going}.
Lower FID values indicate better quality of the generated images.

\noindent
\tb{LPIPS. }
To evaluate diversity, we employ LPIPS~\cite{zhang2018unreasonable} following ~\cite{huang2018multimodal,lee2018diverse,zhu2017toward}.
LIPIS measures the average feature distances between generated samples.
Higher LPIPS score indicates better diversity among the generated images.

\noindent
\tb{NDB and JSD. }
To measure the similarity between the distribution between real images and generated one, we
adopt two bin-based metrics, NDB and JSD, proposed in~\cite{richardson2018NDB}.
These metrics evaluate the extent of mode missing of generative models.
Following~\cite{richardson2018NDB}, the training samples are first clustered using K-means into different bins which can be viewed as modes of the real data distribution.
Then each generated sample is assigned to the bin of its nearest neighbor.
We calculate the bin-proportions of the training samples and the synthesized samples to evaluate the difference between the generated distribution and the real data distribution.
NDB score and JSD of the bin-proportion are then computed to measure the mode collapse.  
Lower NDB score and JSD mean the generated data distribution approaches the real data distribution better by fitting more modes.
Please refer to~\cite{richardson2018NDB} for more details.
\subsection{Conditioned on Class Label}
We first validate the proposed method on categorical generation.
In categorical generation, networks take class labels as conditional contexts to synthesize images of different categories.
We apply the regularization term to the baseline framework DCGAN~\cite{radford2015unsupervised}.

We conduct experiments on the CIFAR-10~\cite{krizhevsky2009learning} dataset which includes images of ten categories.
Since images in the CIFAR-10 dataset are of size $32\times32$ and upsampling degrades the image quality, we do not compute LPIPS in this task.
Table~\ref{tab:NDB_cifar} and Table~\ref{tab:FID_cifar} present the results of NDB, JS, and FID.
MSGAN mitigates the mode collapse issue in most classes while maintaining image quality.
\newcommand{\ra}[1]{\renewcommand{\arraystretch}{#1}}
\renewcommand{\arraystretch}{1.1}
\begin{table}[!h]
	\centering
	\vspace{-1mm}
	\caption{\textbf{FID results on the CIFAR-10 dataset.}}
	\begin{tabular}{@{}ccc@{}} 
	    \toprule 
		Model & DCGAN & MSGAN\\
		\midrule
		FID$\downarrow$ & $29.65\pm{0.06}$ & $\mathbf{28.73\pm{0.06}}$ \\
		\bottomrule 
	\end{tabular}
	\label{tab:FID_cifar}
	\vspace{-3mm}
\end{table}
%
\begin{figure}[h]
\begin{center}
\includegraphics[width=1\linewidth]{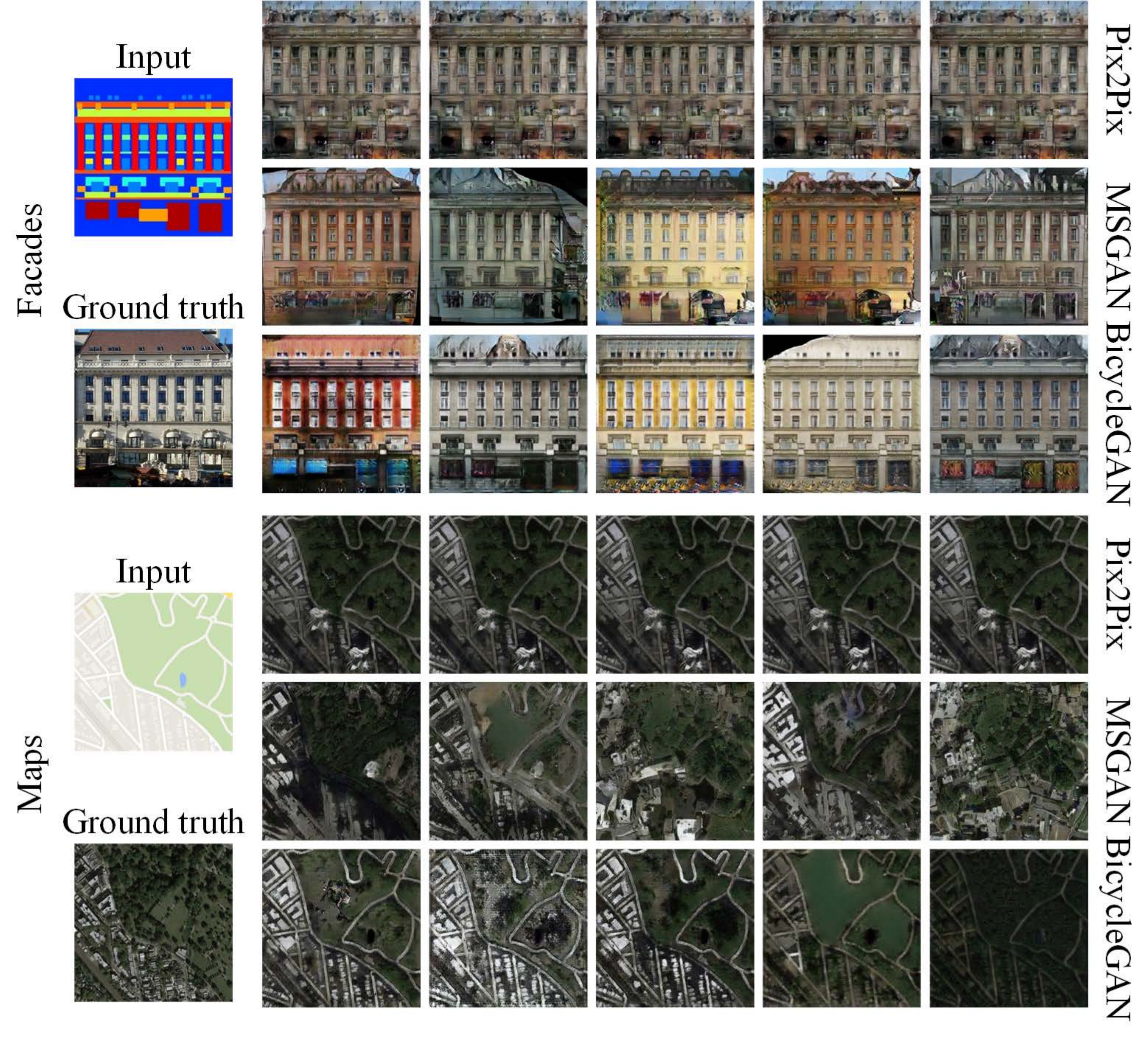}
\end{center}
\vspace{-1mm}
\caption{\textbf{Diversity comparison.} The proposed regularization term helps Pix2Pix learn more diverse results.}
\label{fig:Pix2Pix}
\vspace{-4mm}
\end{figure}

\begin{table}[h]
	\centering
	\caption{\textbf{Quantitative results on the facades and maps dataset.}}
	\small
	\renewcommand{\tabcolsep}{2.5pt}
\vspace{-1mm}
	\begin{tabular}{@{}cccc} 
	    \toprule
	    Datasets &\multicolumn{3}{c}{Facades}\\\cline{2-4}
		 & Pix2Pix~\cite{isola2017image} & MSGAN & BicycleGAN~\cite{zhu2017toward}
		\\\cmidrule{2-4}
		FID $\downarrow$ &$139.19\pm{2.94}$ &$\mathbf{92.84\pm{1.00}}$ &$98.85\pm{1.21}$  \\ 
		NDB$\downarrow$ &$14.40\pm{1.82}$ &$\mathbf{12.40 \pm{0.55}}$&$13.80\pm{0.45}$\\
		JSD$\downarrow$ &$0.074\pm{0.012}$&$\mathbf{0.038\pm{0.004}}$&$0.058\pm{0.004}$\\
		LPIPS$\uparrow$&$0.0003\pm{0.0000}$ &$\mathbf{0.1894\pm{0.0011}}$ & $0.1413\pm{0.0005}$\\
		\midrule
		Datasets &\multicolumn{3}{c}{Maps}\\\cline{2-4}
		  & Pix2Pix~\cite{isola2017image} & MSGAN & BicycleGAN~\cite{zhu2017toward}
		\\\cmidrule{2-4}
		FID $\downarrow$ &$168.99\pm{2.58}$ &$152.43\pm{2.52}$ &$\mathbf{145.78\pm{3.90}}$  \\ 
		NDB$\downarrow$ &$49.00\pm{1.00}$ &$\mathbf{41.60 \pm{0.55}}$&$46.60\pm{1.34}$\\
		JSD$\downarrow$ &$0.088\pm{0.018}$&$0.031\pm{0.003}$&$\mathbf{0.023\pm{0.002}}$\\
		LPIPS$\uparrow$&$0.0016\pm{0.0003}$ &$\mathbf{0.2189\pm{0.0004}}$ & $0.1150\pm{0.0007}$\\
		\bottomrule
	\end{tabular}
	\vspace{-4mm}
	\label{tab:label2photos}
\end{table}

\begin{figure*}[t]
\begin{center}
\includegraphics[width=1\linewidth]{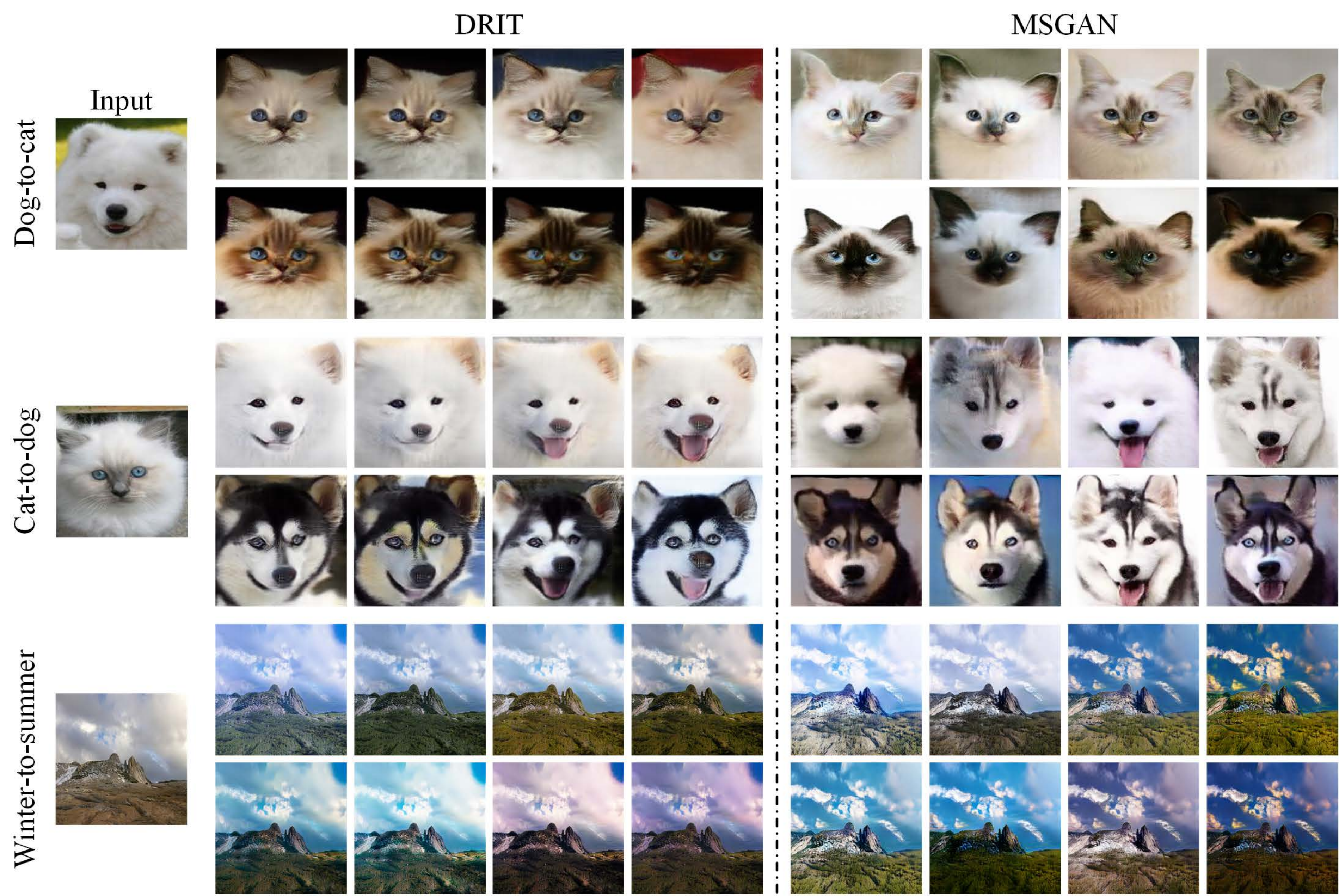}
\end{center}
\vspace{-1mm}
\caption{\textbf{Diversity comparison.}
We compare MSGAN with DRIT on the dog-to-cat, cat-to-dog, and winter-to-summer translation tasks.
Our model produces more diverse samples over DRIT.}
\vspace{-3mm}
\label{fig:DRIT}
\end{figure*}

\begin{figure}[h]
\begin{center}
\includegraphics[width=1\linewidth]{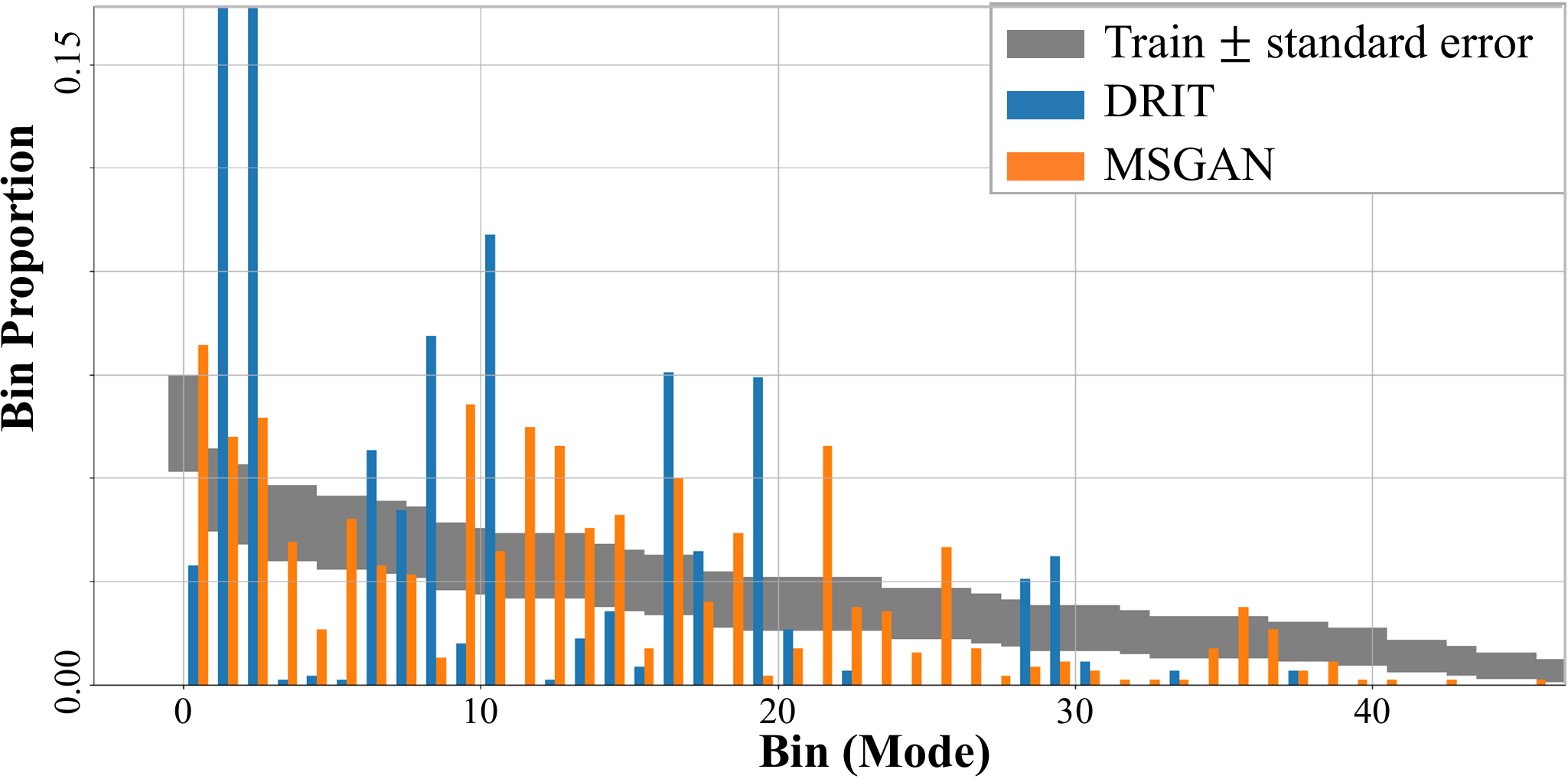}
\end{center}
\vspace{-3mm}
\caption{\textbf{Visualization of the bins on dog$\rightarrow$cat translation.} 
The translated results of DRIT collapse into few modes, while the generated image of MSGAN fit the real data distribution better.}
\label{fig:NDB}
\vspace{-3mm}
\end{figure}

\subsection{Conditioned on Image}
Image-to-image translation aims to learn the mapping between two visual domains.
Conditioned on images from the source domain, models attempt to synthesize corresponding images in the target domain.
Despite the multimodal nature of the image-to-image translation task, early work~\cite{isola2017image, zhu2017unpaired} abandons noise vectors and performs one-to-one mapping since the latent codes are easily ignored during training as shown in~\cite{isola2017image,zhu2017toward}.
To achieve multimodality, several recent attempts~\cite{huang2018multimodal,lee2018diverse,zhu2017toward} introduce additional encoder networks and objective functions to impose a bijection constraint between the latent code space and the image space. 
To demonstrate the generalizability, we apply the proposed method to a unimodal model Pix2Pix~\cite{isola2017image} using paired training data and a multimodal model DRIT~\cite{lee2018diverse} using unpaired images.

\begin{figure*}[t]
\begin{center}
\includegraphics[width=1\linewidth]{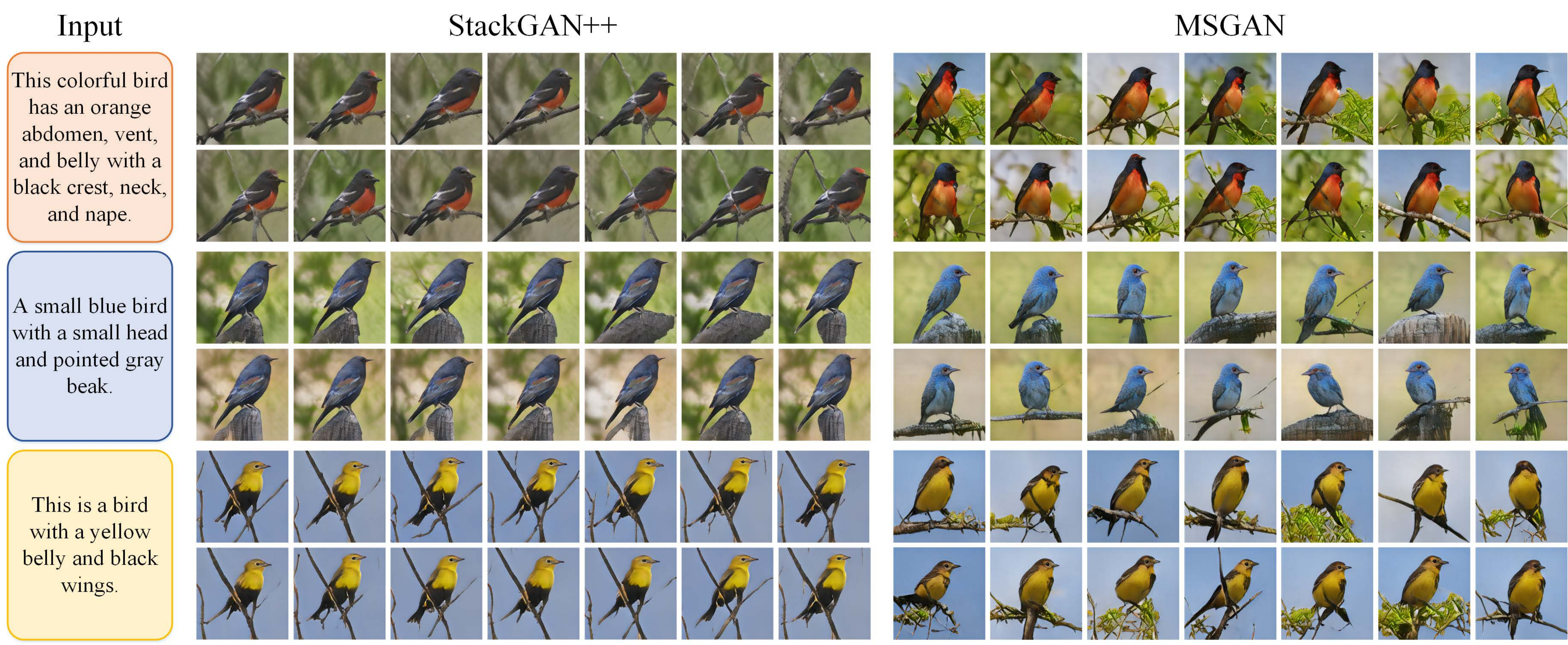}
\end{center}
\vspace{-1mm}
\caption{\textbf{Diversity comparison.} We show examples of StackGAN++~\cite{zhang2017stackgan++} and MSGAN on the CUB-200-2011 dataset of text-to-image synthesis.
When the text code is fixed, the latent codes in MSGAN help to generate more diverse appearances and poses of birds as well as different backgrounds.}
\vspace{-1mm}
\label{fig:stackGAN}
\end{figure*}

\begin{table*}[t]
	\centering
	\caption{\textbf{Quantitative results of the Yosemite (Summer$\rightleftharpoons$Winter) and the Cat$\rightleftharpoons$Dog dataset.}}
	\begin{tabular}{@{}ccc ccc} 
	    \toprule
		Datasets & \multicolumn{2}{c}{Summer2Winter}&\multicolumn{2}{c}{Winter2Summer}
		\\  \cmidrule(lr){2-3} \cmidrule(lr){4-5}
		 & DRIT~\cite{lee2018diverse} & MSGAN & DRIT~\cite{lee2018diverse} & MSGAN \\
		FID $\downarrow$ &$57.24\pm{2.03}$& $\mathbf{51.85\pm{1.16}}$ & $47.37\pm{3.25}$ &$\mathbf{46.23\pm{2.45}}$ \\ 
		NDB$\downarrow$ & $25.60\pm{1.14}$ & $\mathbf{22.80\pm{2.96}}$&  $30.60\pm{2.97}$ & $\mathbf{27.80\pm{3.03}}$ \\
		JSD$ \downarrow$ &$0.066\pm{0.005}$&$\mathbf{0.046\pm{0.006}}$&$0.049\pm{0.009}$&$\mathbf{0.038\pm{0.004}}$\\
		LPIPS$\uparrow$&$0.1150\pm{0.0003}$& $\mathbf{0.1468\pm{0.0005}}$& $0.0965\pm{0.0004}$& $\mathbf{0.1183\pm{0.0007}}$
		\\ \midrule
		Datasets& \multicolumn{2}{c}{Cat2Dog} & \multicolumn{2}{c}{Dog2Cat}\\ \cmidrule(lr){2-3} \cmidrule(lr){4-5}
		& DRIT~\cite{lee2018diverse} & MSGAN & DRIT~\cite{lee2018diverse} & MSGAN \\
		FID$\downarrow$ &$22.74\pm{0.28}$ & $\mathbf{16.02\pm{0.30}}$ & $62.85\pm{0.21}$ & $\mathbf{29.57\pm{0.23}}$\\ 
		NDB$\downarrow$ & $42.00\pm{2.12}$ &$\mathbf{27.20\pm{0.84}}$ &   $41.00\pm{0.71}$ & $\mathbf{31.00\pm{0.71}}$  \\
		JSD$ \downarrow$ &$0.127\pm{0.003}$& $\mathbf{0.084\pm{0.002}}$ &$0.272\pm{0.002}$ &$\mathbf{0.068\pm{0.001}}$ \\
		LPIPS$\uparrow$&$0.245\pm{0.002}$& $\mathbf{0.280\pm{0.002}}$ & $0.102\pm{0.001}$ & $\mathbf{0.214\pm{0.001}}$\\
		\bottomrule 
	\end{tabular}
	\label{tab:drit}
	\vspace{-1mm}
\end{table*}

\begin{table*}[h]
\centering
\caption{\textbf{Quantitative results on the CUB-200-2011 dataset.}
We conduct experiments in two settings:
1) Conditioned on text descriptions, where every description can be mapped to different text codes.
2) Conditioned on text codes, where the text codes are fixed so that their effects are excluded.
}
\vspace{-1mm}
\begin{tabular}{@{}ccccc} 
\toprule
	  &\multicolumn{2}{c}{Conditioned on text descriptions} & \multicolumn{2}{c}{Conditioned on text codes}\\
	       \cmidrule(lr){2-3} \cmidrule(lr){4-5}
& StackGAN++~\cite{zhang2017stackgan++} & MSGAN & StackGAN++~\cite{zhang2017stackgan++} & MSGAN  \\
FID $\downarrow$ &$25.99\pm{4.26}$ &$\mathbf{25.53\pm{1.83}}$ & $\mathbf{27.12\pm{1.15}}$ &$27.94\pm{3.10}$ \\ 	NDB$\downarrow$ & $38.20\pm{2.39}$ &$\mathbf{30.60\pm{2.51}}$&$39.00\pm{0.71}$&$\mathbf{30.60\pm{2.41}}$ \\
JSD$\downarrow$ & $0.092\pm{0.005}$&$\mathbf{0.073\pm{0.003}}$&$0.102\pm{0.016}$  & $\mathbf{0.095\pm{0.016}}$ \\
LPIPS$\uparrow$&$0.362\pm{0.004}$ & $\mathbf{0.373\pm{0.007}}$ &$0.156\pm{0.004}$ &  $\mathbf{0.207\pm{0.005}}$\\
\bottomrule 	
\end{tabular}
\label{tab:CUB}
\vspace{-1mm}
\end{table*}

\begin{figure*}[h]
\begin{center}
\includegraphics[width=1\linewidth]{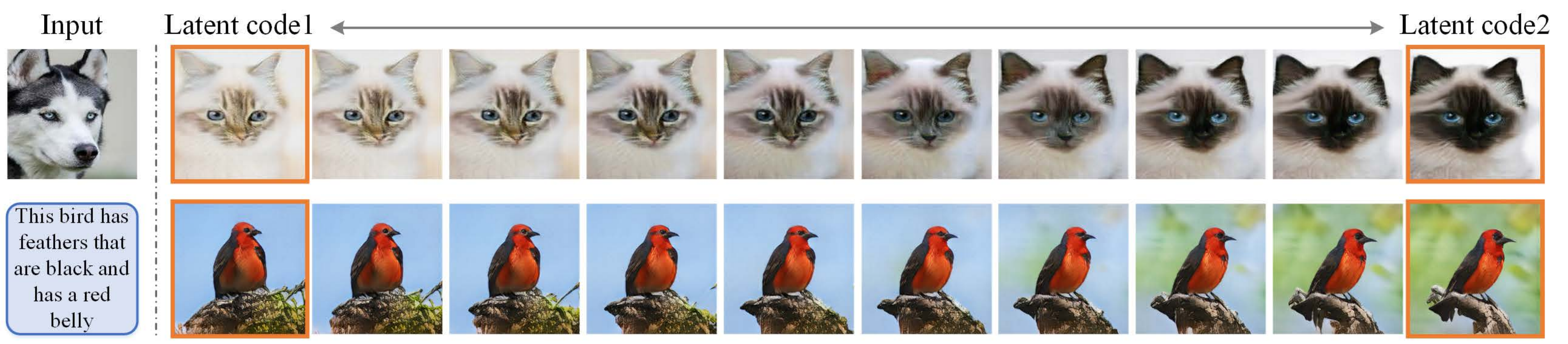}
\end{center}
\vspace{-4mm}
\caption{\textbf{Linear interpolation between two latent codes in MSGAN.} Image synthesis results with linear-interpolation between two latent codes in the dog-to-cat translation and text-to-image synthesis.}
\label{fig:interpolation}
\vspace{-4mm}
\end{figure*}

\subsubsection{Conditioned on Paired Images}
\label{pair}
We take Pix2Pix as the baseline model.
We also compare MSGAN to BicycleGAN~\cite{zhu2017toward} which generates diverse images with paired training images.
For fair comparisons, architectures of the generator and the discriminator in all methods follow the ones in BicycleGAN~\cite{zhu2017toward}.

We conduct experiments on the facades and maps datasets.
MSGAN obtains consistent improvements on all metrics over Pix2Pix.
Moreover, MSGAN demonstrates comparable diversity to BicycleGAN, which applies an additional encoder network.
Figure.~\ref{fig:Pix2Pix} and Table.~\ref{tab:label2photos} demonstrate the qualitative and quantitative results, respectively.
\vspace{-7mm}
\subsubsection{Conditioned on Unpaired Images}
\label{unpaired}
We choose DRIT~\cite{lee2018diverse}, one of the state-of-the-art frameworks to generate diverse images with unpaired training data, as the baseline framework.
Though DRIT synthesizes diverse images in most cases, mode collapse occurs in some challenging shape-variation cases (\eg translation between cats and dogs).
To demonstrate the robustness of the proposed method, we evaluate on the shape-preserving Yosemite (summer$\rightleftharpoons$winter)~\cite{zhu2017unpaired} dataset and the cat$\rightleftharpoons$dog~\cite{lee2018diverse} dataset that requires shape variations. 

As the quantitative results exhibited in Table.~\ref{tab:drit}, MSGAN performs favorably against DRIT in all metrics on both datasets.
Especially in the challenging cat$\rightleftharpoons$dog dataset, MSGAN obtains substantial diversity gains.
From the statistical point of view, we visualize the bin proportions of the dog-to-cat translation in Figure.~\ref{fig:NDB}. 
The graph shows the severe mode collapse issue of DRIT and the substantial improvement with the proposed regularization term.
Qualitatively, Figure.~\ref{fig:DRIT} shows that MSGAN discovers more modes without the loss of visual quality.

\subsection{Conditioned on Text}
\vspace{-5mm}
Text-to-image synthesis targets at generating images conditioned on text descriptions.
We integrate the proposed regularization term on StackGAN++~\cite{zhang2017stackgan++} using the CUB-200-2011~\cite{WahCUB_200_2011} dataset.
To improve diversity, StackGAN++ introduces a Conditioning Augmentation~(CA) module that re-parameterizes text descriptions into text codes of the Gaussian distribution.
Instead of applying the regularization term on the semantically meaningful text codes, we focus on exploiting the latent codes randomly sampled from the prior distribution.
However, for a fair comparison, we evaluation MSGAN against StackGAN++ in two settings:
1) Perform generation without fixing text codes for text descriptions. In this case, text codes also provide variations for output images.
2) Perform generation with fixed text codes. In this setting, the effects of text codes are excluded.

Table.~\ref{tab:CUB} presents quantitative comparisons between MSGAN and StackGAN++.
MSGAN improves the diversity of StackGAN++ and maintains visual quality.
To better illustrate the role that latent codes play for the diversity, we show qualitative comparisons with the text codes fixed.
In this setting, we do not consider the diversity resulting from CA.
Figure.~\ref{fig:stackGAN} illustrates that latent codes of StackGAN++ have minor effects on the variations of the image.
On the contrary, latent codes of MSGAN contribute to various appearances and poses of birds.

\subsection{Interpolation of Latent Space in MSGANs}

We perform linear interpolation between two given latent codes and generate corresponding images to have a better understanding of how well MSGANs exploit the latent space.
Figure.~\ref{fig:interpolation} shows the interpolation results on the dog-to-cat translation and the text-to-image synthesis task.
In the dog-to-cat translation, we can see the coat colors and patterns varies smoothly along with the latent vectors.
In the text-to-image synthesis, both orientations of birds and the appearances of footholds change gradually with the variations of the latent codes.

\vspace{-2mm}
\section{Conclusions}
\vspace{-1mm}
In this work, we present a simple but effective mode seeking regularization term on the generator to address the model collapse in cGANs.
By maximizing the distance between generated images with respect to that between the corresponding latent codes, the regularization term forces the generators to explore more minor modes.
The proposed regularization method can be readily integrated with existing cGANs framework without imposing training overheads and modifications of network structures.
We demonstrate the generalizability of the proposed method on three different conditional generation tasks including categorical generation, image-to-image translation, and text-to-image synthesis.
Both qualitative and quantitative results show that the proposed regularization term facilitates the baseline frameworks improving the diversity without sacrificing visual quality of the generated images.

\clearpage
{\small
\bibliographystyle{ieee}
\bibliography{egbib}
}
\clearpage
\begin{appendices}
\section{Implementation Details}
Table~\ref{tab:datasets} summarizes the datasets and baseline models used on various tasks. 
For all of the baseline methods, we incorporate the original objective functions with the proposed regularization term.
Note that we remain the original network architecture design and use the default setting of hyper-parameters for the training.
\Paragraph{DCGAN.} 
Since the images in the CIFAR-10~\cite{krizhevsky2009learning} dataset are of size $32\times32$, we modify the structure of the generator and discriminator in DCGAN~\cite{radford2015unsupervised}, as shown in Table~\ref{tab:DCGAN}.
We use the batch size of $32$, learning rate of $0.0002$ and Adam~\cite{kingma2014adam} optimizer with $\beta_1=0.5$ and $\beta_2=0.999$ to train both the baseline and MSGAN network.

\Paragraph{Pix2Pix.} We adopt the generator and discriminator in BicycleGAN~\cite{zhu2017unpaired} to build the Pix2Pix~\cite{isola2017image} model.
Same as BicycleGAN, we use a U-Net network~\cite{ronneberger2015u} for the generator, and inject the latent codes $\mathbf{z}$ into every layer of the generator.
The architecture of the discriminator is a two-scale PatchGAN network~\cite{isola2017image}.
For the training, both Pix2Pix and MSGAN framework use the same hyper-parameters as the officially released version~\footnote{\url{https://github.com/junyanz/BicycleGAN/}}.

\Paragraph{DRIT.}
DRIT~\cite{lee2018diverse} involves two stages of image-to-image translations in the training process.
We only apply the mode seeking regularization term to generators in the first stage, which is modified on the officially released code~\footnote{\url{https://github.com/HsinYingLee/DRIT}}.

\Paragraph{StackGAN++.}
StackGAN++~\cite{zhang2017stackgan++} is a tree-like structure with multiple generators and discriminators.
We use the output images from the last generator and input latent codes to calculate the mode seeking regularization term.
The implementation is based on the officially released code~\footnote{\url{https://github.com/hanzhanggit/StackGAN-v2}}.

\section{Evaluation Details}
%
We employ the official implementation of FID~\footnote{\url{https://github.com/bioinf-jku/TTUR}}, NDB and JSD~\footnote{\url{https://github.com/eitanrich/gans-n-gmms}}, and LPIPS~\footnote{\url{https://github.com/richzhang/PerceptualSimilarity}}.
For NDB and JSD, we use the K-means method on training samples to obtain the clusters.
Then the generated samples are assigned to the nearest cluster to compute the bin proportions. 
As suggested by the author of~\cite{richardson2018NDB}, there are at least $10$ training samples for each cluster.
Therefore, we cluster the number of bins~$K \approx N_{\rm train}/{20}$ in all tasks, where $N_{\rm train}$ denotes the number of training samples for computing the clusters.
We have verified that the performance is consistent within a large range of $K$.
For evaluation, we randomly generate $N$ images for a given conditional context on various tasks.
We conduct five independent trials and report the mean and standard derivation based on the result of each trial.
More evaluation details of one trial are presented as follows.
\Paragraph{Conditioned on Class Label.} We randomly generate $N=5000$ images for each class label.
We use all the training samples and the generated samples to compute FID.
For NDB and JSD, we employ the training samples in each class to calculate $K=250$ clusters.

\Paragraph{Conditioned on Image.} We randomly generate $N=50$ images for each input image in the test set.
For LPIPS, we randomly select $50$ pairs of the $50$ images of each context in the test set to compute LPIPS and average all the values for this trial.
Then, we randomly choose $100$ input images and their corresponding generated images to form $5000$ generated samples.
We use the $5000$ generated samples and all samples in training set to compute FID.
For NDB and JSD, we employ all the training samples for clustering and choose $K=20$ bins for facades, and $K=50$ bins for other datasets.
\Paragraph{Conditioned on Text.} We randomly select $200$ sentences and generate $N=10$ images for each sentence, which forms $2000$ generated samples.
Then, we randomly select $N_{\rm train}= 2000$ samples for computing FID, and clustering them into $K=100$ bins for NDB and JSD.
For LPIPS, we randomly choose $10$ pairs for each sentence and average the values of all the pairs for this trial.

\section{Ablation Study on the Regularization Term}
\subsection{The Weighting Parameter $\lambda_{\mathrm{ms}}$}
To analyze the influence of the regularization term, we conduct an ablation study by varying the weighting parameter $\lambda_{\mathrm{ms}}$ on image-to-image translation task using the facades dataset.
Table.~\ref{tab:lambda} presents the quantitative results with diverse $\lambda_{\mathrm{ms}}$.
It can be observed that increasing $\lambda_{\mathrm{ms}}$ improves the diversity of the generated images.
Nevertheless, as the weighting parameter $\lambda_{\mathrm{ms}}$ becomes larger than a threshold value ($1.0$), the training becomes unstable, which yields low quality, and even low diversity synthesized images.
As a result, we empirically set the weighting parameter $\lambda_{\mathrm{ms}}=1.0$ for all experiments.
\begin{table}[!h]
	\centering
	\caption{\textbf{Quantitative results with different $\lambda_{\mathrm{ms}}$ on the facades dataset.}}
	\renewcommand{\tabcolsep}{2.5pt}
\vspace{-1mm}
	\begin{tabular}{@{}c|ccccc} 
	    \toprule
		 $\lambda_{\rm ms}$& $0$ & $0.1$ & $0.5$ & $1$ & $3$
		\\\midrule
		FID $\downarrow$ &$134.95$&$126.31$ &$88.41$ &$92.45$& $338.45$  \\ 
		LPIPS$\uparrow$&$0.0003$&$0.0155$ &$0.0929$ & $\mathbf{0.1888}$ &$0.1393$ \\
		\bottomrule
	\end{tabular}
	\vspace{-3mm}
	\label{tab:lambda}
\end{table}
\subsection{The Design Choice of the Distance Metric}
We have explored other design choice of the distance metric.
We conduct experiments using discriminator feature distance in our regularization term in a way similar to feature matching loss~\cite{wang2018pix2pixHD}, 
\begin{equation}
    \mathcal{L}_{ms} = \frac{\frac{1}{L}\sum_{l=1}^{L}\|D^{l}(G(\mathbf{c},\mathbf{z}_2))-D^{l}(G(\mathbf{c},\mathbf{z}_1))\|_1}{\|\mathbf{z}_2-\mathbf{z}_1\|_1},
\end{equation}
where $D^l$ denotes the $l^{th}$ layer of the discriminator. 
We apply it to Pix2Pix on the facades dataset.
Table.~\ref{tab:distance-choice} shows that MSGAN using feature distance also obtains improvement over Pix2Pix.
However, MSGAN using $L_1$ distance has higher diversity.
Therefore, we employ MSGAN using $L_1$ distance for all experiments.
\begin{table}[!h]
	\centering
	\caption{\textbf{Quantitative results on the facades dataset.}}
	\small
	\renewcommand{\tabcolsep}{2.5pt}
\vspace{-1mm}
	\begin{tabular}{@{}cccc} 
	    \toprule
		 & Pix2Pix~\cite{isola2017image} & MSGAN-L$_1$ & MSGAN-FD
		\\\midrule
		FID $\downarrow$ &$139.19\pm{2.94}$ &$\mathbf{92.84\pm{1.00}}$ &$100.16\pm{3.14}$  \\ 
		NDB$\downarrow$ &$14.40\pm{1.82}$ &$12.40 \pm{0.55}$&$\mathbf{11.80\pm{1.48}}$\\
		JSD$\downarrow$ &$0.074\pm{0.012}$&$\mathbf{0.038\pm{0.004}}$&$0.072\pm{0.014}$\\
		LPIPS$\uparrow$&$0.0003\pm{0.0000}$ &$\mathbf{0.1894\pm{0.0011}}$ & $0.0565\pm{0.0003}$ \\
		\bottomrule
	\end{tabular}
	\vspace{-3mm}
	\label{tab:distance-choice}
\end{table}

\section{Computational Overheads}
We compare MSGAN with Pix2Pix, BicycleGAN in terms of training time, memory consumption, and model parameters on an NVIDIA TITAN X GPU.
Table.~\ref{tab:overheads} shows that our method incurs marginal overheads.
However, BicycleGAN requires longer time per iteration and larger memory with an additional encoder and another discriminator network.
\begin{table}[!h]
	\centering
	\small
		\caption{\textbf{Comparisons of computational overheads on the facades dataset.}}
	\begin{tabular}{@{}cccc@{}} 
	    \toprule 
		Model & Time (s) & Memory (MB) & Parameters (M)\\
		\midrule
		Pix2Pix~\cite{isola2017image} & $0.122$ &$1738$  & $58.254$\\
		MSGAN & $0.122$ &$1739$ & $58.254$\\
		BicycleGAN~\cite{zhu2017toward} &$0.192$  & $2083$ &$64.303$\\
		\bottomrule 
	\end{tabular}
	\label{tab:overheads}
\end{table}

\section{Additional Results}
\begin{figure*}[h]
\begin{center}
\includegraphics[width=1\linewidth]{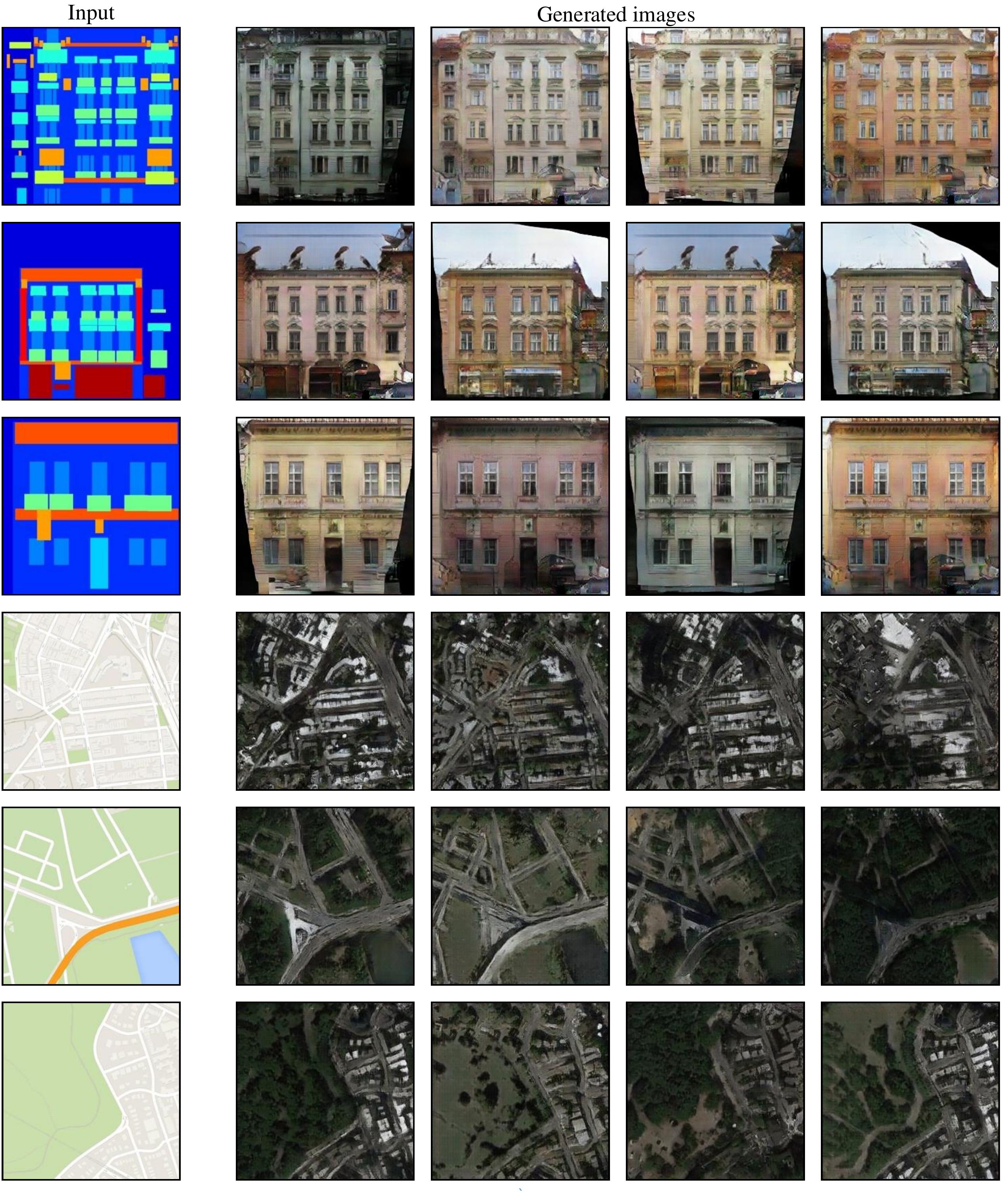}
\end{center}
\vspace{-1mm}
\caption{\textbf{More image-to-image translation results of facades and maps.} Top three rows: facades, bottom three rows: maps.}
\label{fig:facades}
\vspace{-3mm}
\end{figure*}

\begin{figure*}[h]
\begin{center}
\includegraphics[width=1\linewidth]{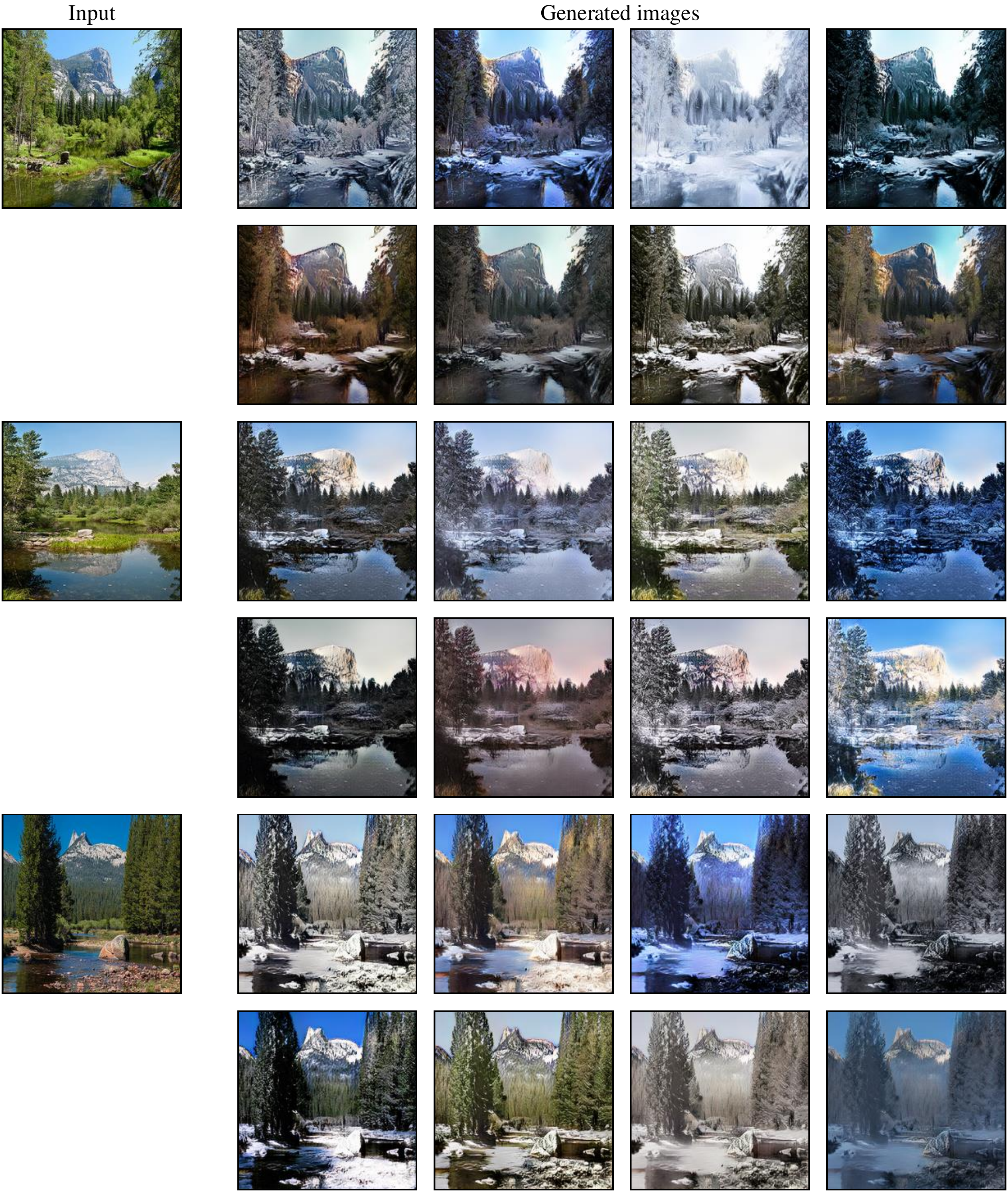}
\end{center}
\vspace{-1mm}
\caption{\textbf{More image-to-image translation results of Yosemite, Summer$\rightarrow$Winter.}}
\label{fig:summer2winter}
\vspace{-3mm}
\end{figure*}

\begin{figure*}[h]
\begin{center}
\includegraphics[width=1\linewidth]{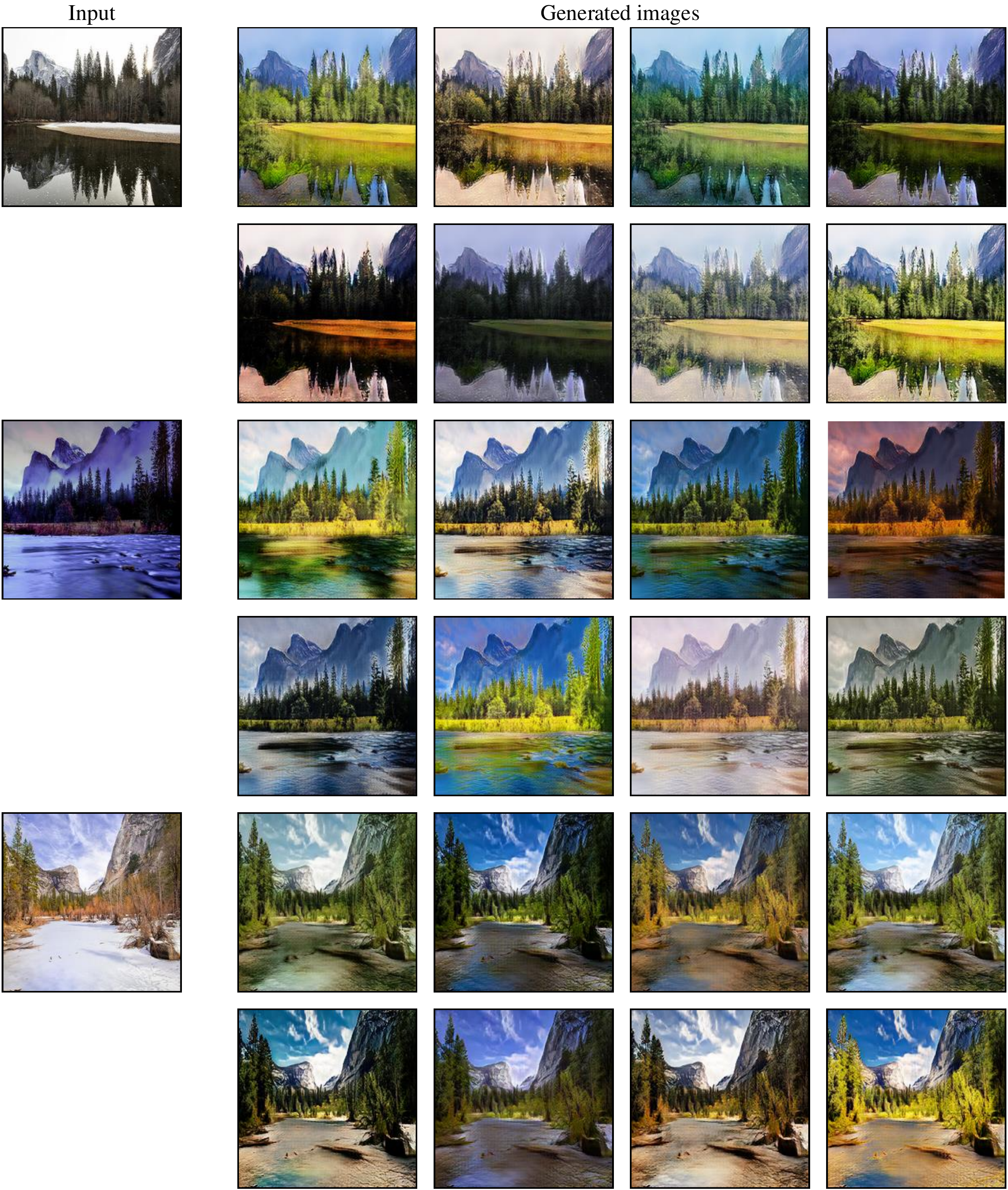}
\end{center}
\vspace{-1mm}
\caption{\textbf{More image-to-image translation results of Yosemite, Winter$\rightarrow$Summer.}}
\label{fig:winter2summer}
\vspace{-3mm}
\end{figure*}

\begin{figure*}[h]
\begin{center}
\includegraphics[width=1\linewidth]{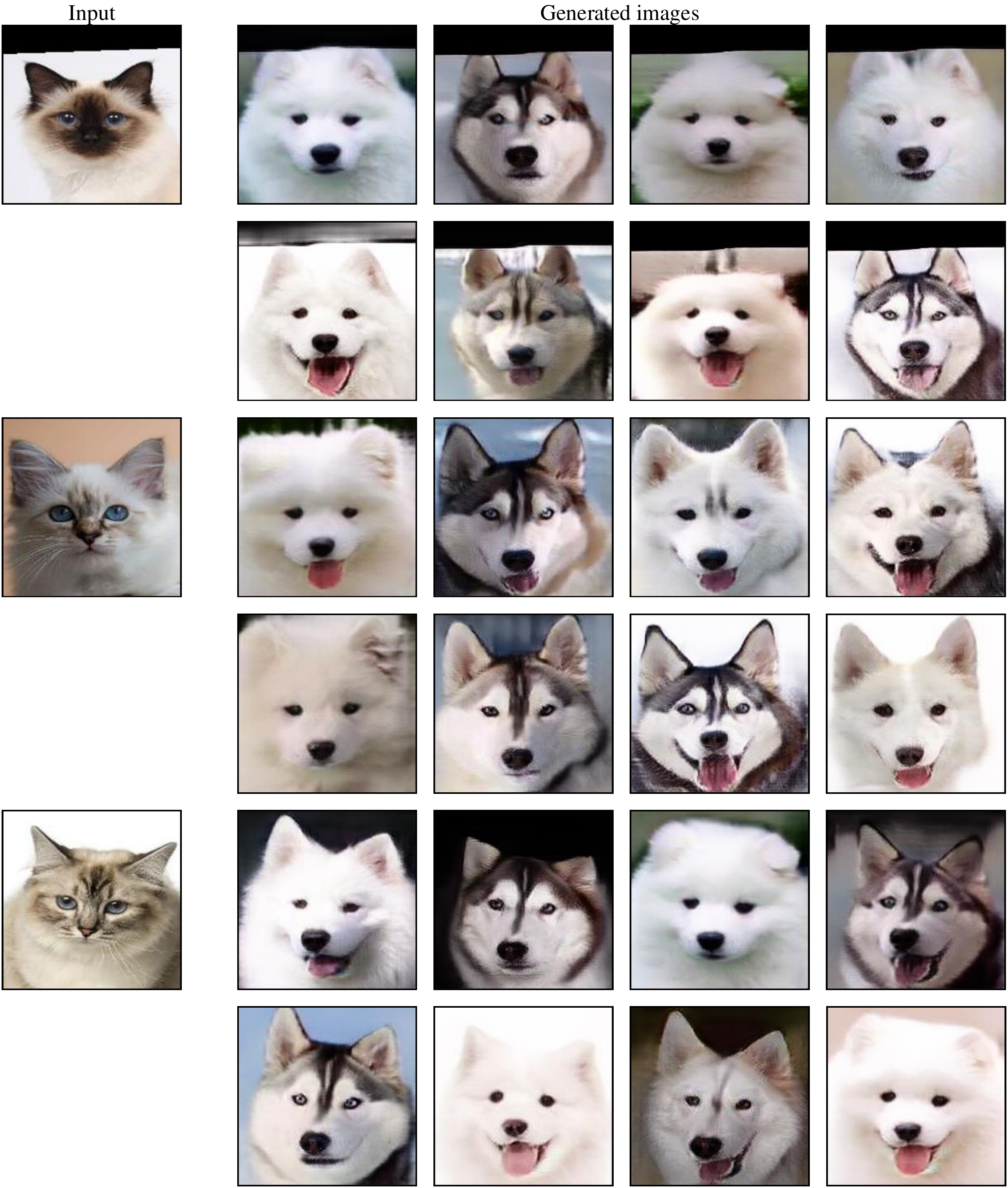}
\end{center}
\vspace{-1mm}
\caption{\textbf{More image-to-image translation results of Cat$\rightarrow$Dog.}}
\label{fig:cat2dog}
\vspace{-3mm}
\end{figure*}

\begin{figure*}[h]
\begin{center}
\includegraphics[width=1\linewidth]{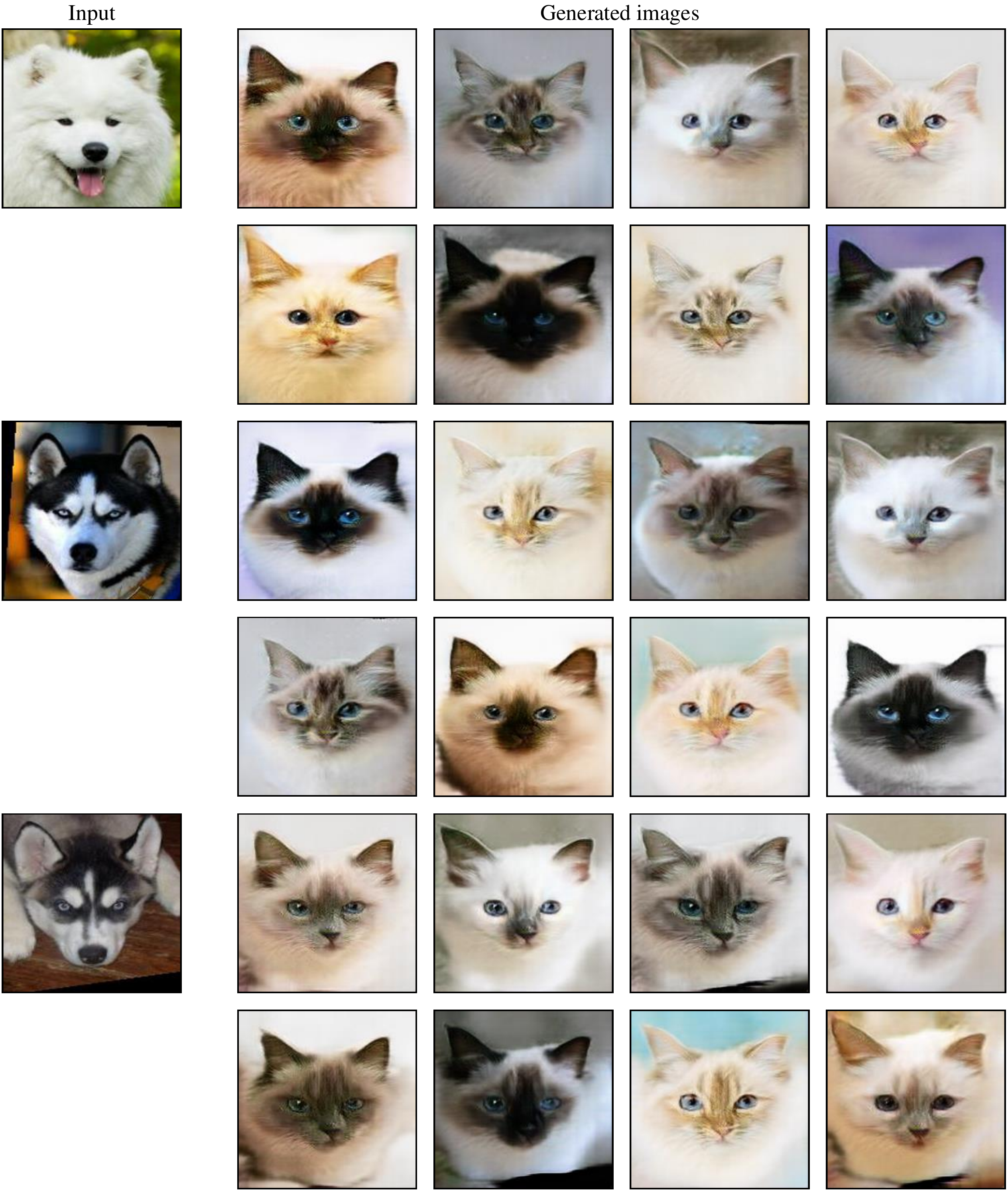}
\end{center}
\vspace{-1mm}
\caption{\textbf{More image-to-image translation results of Dog$\rightarrow$Cat.}}
\label{fig:dog2cat}
\vspace{-3mm}
\end{figure*}

\begin{figure*}[h]
\begin{center}
\includegraphics[width=1\linewidth]{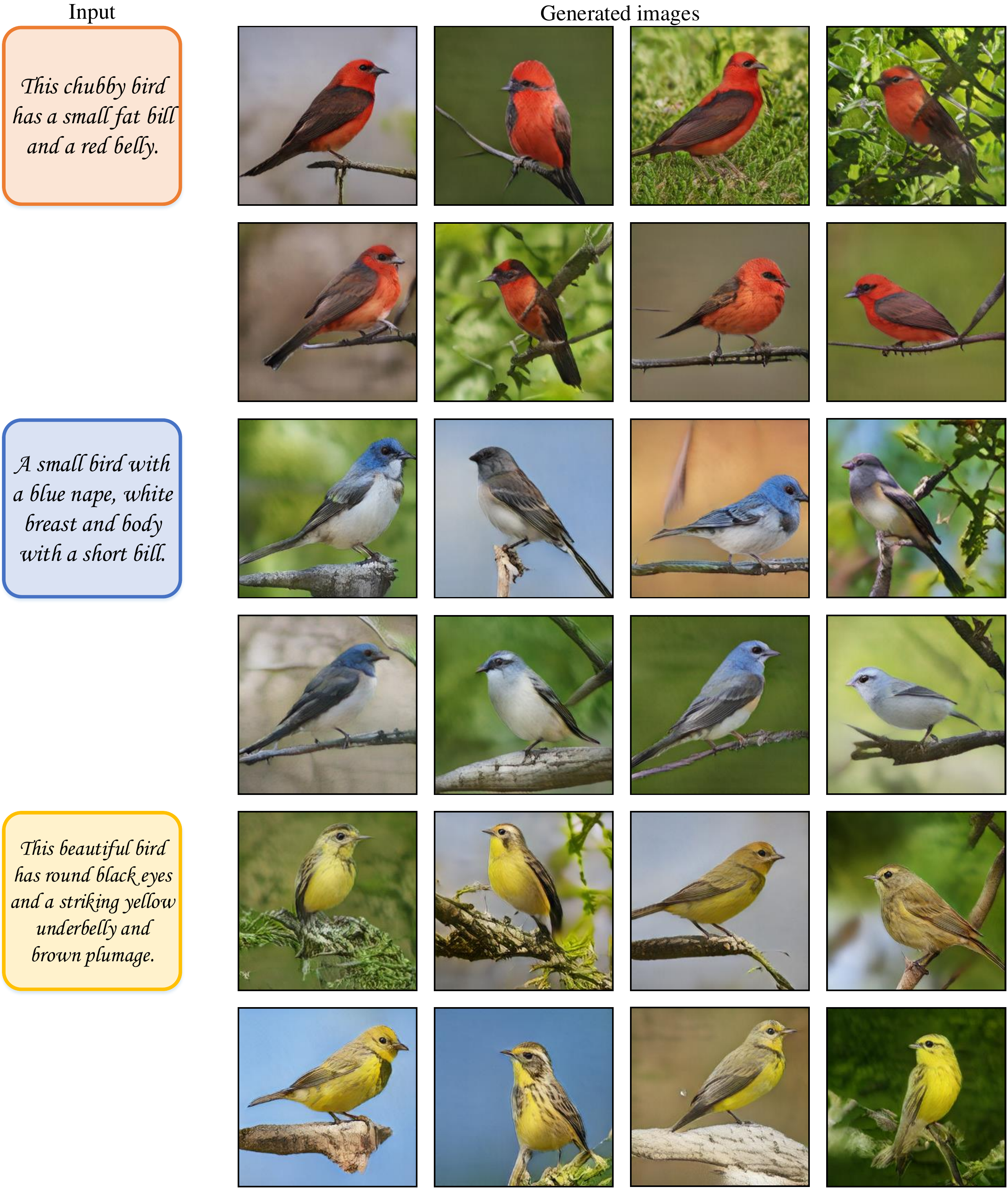}
\end{center}
\vspace{-1mm}
\caption{\textbf{More text-to-image synthesis results of CUB-200-2011.}}
\label{fig:bird}
\vspace{-3mm}
\end{figure*}

We present more results of categorical generation, image-to-image translation, and text-to-image synthesis in Figure~\ref{fig:facades}, Figure~\ref{fig:summer2winter}, Figure~\ref{fig:winter2summer}, Figure~\ref{fig:cat2dog}, Figure~\ref{fig:dog2cat}, and Figure~\ref{fig:bird}, respectively.
\begin{table*}[h]
\footnotesize
\centering
	\caption{\textbf{Statistics of different generation tasks.} We summarize the number of training and testing images in each generation task. The baseline model used for each task is also provided.}
	\begin{tabular}{l c c c c c c c c c c c c c c c c} 
	  \toprule
	  Context & \multicolumn{2}{c}{Class Label} & \multicolumn{4}{c}{Paired Images}  & \multicolumn{8}{c}{Unpaired Images}&\multicolumn{2}{c}{Text} \\ 		\cmidrule(lr){2-3}\cmidrule(lr){4-7} \cmidrule(lr){8-15} \cmidrule(lr){16-17}
		\multirow{3}{*}{Dataset}& \multicolumn{2}{c}{\multirow{2}{*}{CIFAR-10~\cite{krizhevsky2009learning}}}  &\multicolumn{2}{c}{\multirow{2}{*}{Facades~\cite{cordts2016cityscapes}}} 
		&\multicolumn{2}{c}{\multirow{2}{*}{Maps~\cite{isola2017image}}}
		&\multicolumn{4}{c}{Yosemite~\cite{zhu2017unpaired}} 
		&\multicolumn{4}{c}{Cat $\rightleftharpoons$ Dog~\cite{lee2018diverse}} & \multicolumn{2}{c}{\multirow{2}{*}{CUB-200-2011~\cite{WahCUB_200_2011}}}
		 \\
		 \cmidrule(lr){8-11} \cmidrule(lr){12-15}

		& \multicolumn{2}{c}{}&\multicolumn{2}{c}{}&\multicolumn{2}{c}{}& \multicolumn{2}{c}{Summer}&  \multicolumn{2}{c}{Winter}&\multicolumn{2}{c}{Cat}& \multicolumn{2}{c}{Dog}& \multicolumn{2}{c}{}\\
		\cmidrule(lr){2-3}\cmidrule(lr){4-5}\cmidrule(lr){6-7}\cmidrule(lr){8-9}\cmidrule(lr){10-11}\cmidrule(lr){12-13} \cmidrule(lr){14-15} \cmidrule(lr){16-17}
         
		&train& test &train& test &train & test& train & test &train & test & train & test&train&test & train& test \\
		\cmidrule(lr){2-3}\cmidrule(lr){4-5}\cmidrule(lr){6-7}\cmidrule(lr){8-9}\cmidrule(lr){10-11}\cmidrule(lr){12-13} \cmidrule(lr){14-15} \cmidrule(lr){16-17}
		Samples& $50,000$&$10,000$&$400$& $206$& $1,096$ & $1,098$ & $1,069$ & $309$ & $770$& $238$& $771$ & $100$& $1,264$&$100$ &$8,855$ & $2,933$   \\
		\cmidrule(lr){2-3}\cmidrule(lr){4-7} \cmidrule(lr){8-15} \cmidrule(lr){16-17}
		Baseline&\multicolumn{2}{c}{DCGAN~\cite{radford2015unsupervised}} & \multicolumn{4}{c}{Pix2Pix~\cite{isola2017image}}&\multicolumn{8}{c}{DRIT~\cite{lee2018diverse}}&\multicolumn{2}{c}{StackGAN++~\cite{zhang2017stackgan++}}
		\\
		\bottomrule
	\end{tabular}
	\label{tab:datasets}
\end{table*}

\begin{table*}[h]
\centering
\caption{
	\textbf{The architecture of the generator and discriminator of DCGAN}. We employ the following abbreviation: N= Number of filters, K= Kernel size, S= Stride size, P= Padding size.
	``Conv'', ``Dconv'',``BN'' denote the convolutional layer, transposed convolutional layer and batch normalization, respectively.
}
\begin{tabular}{@{}l l l@{}}
		\toprule
		Layer & Generator & Discriminator \\
		\midrule
    $1$ & Dconv(N512-K4-S1-P0), BN, Relu  & Conv(N128-K4-S2-P1), Leaky-Relu  \\
    $2$ & Dconv(N256-K4-S2-P1), BN, Relu  & Conv(N256-K4-S2-P1), BN, Leaky-Relu \\
    $3$ & Dconv(N128-K4-S2-P1), BN, Relu  & Conv(N512-K4-S2-P1), BN, Leaky-Relu\\
    $4$ & Dconv(N3-K4-S2-P1), Tanh &  Conv(N1-K4-S1-P0), Sigmoid \\
		\bottomrule
\end{tabular}
\label{tab:DCGAN}
\end{table*}


\begin{figure*}[h]
\begin{center}
\includegraphics[width=1\linewidth]{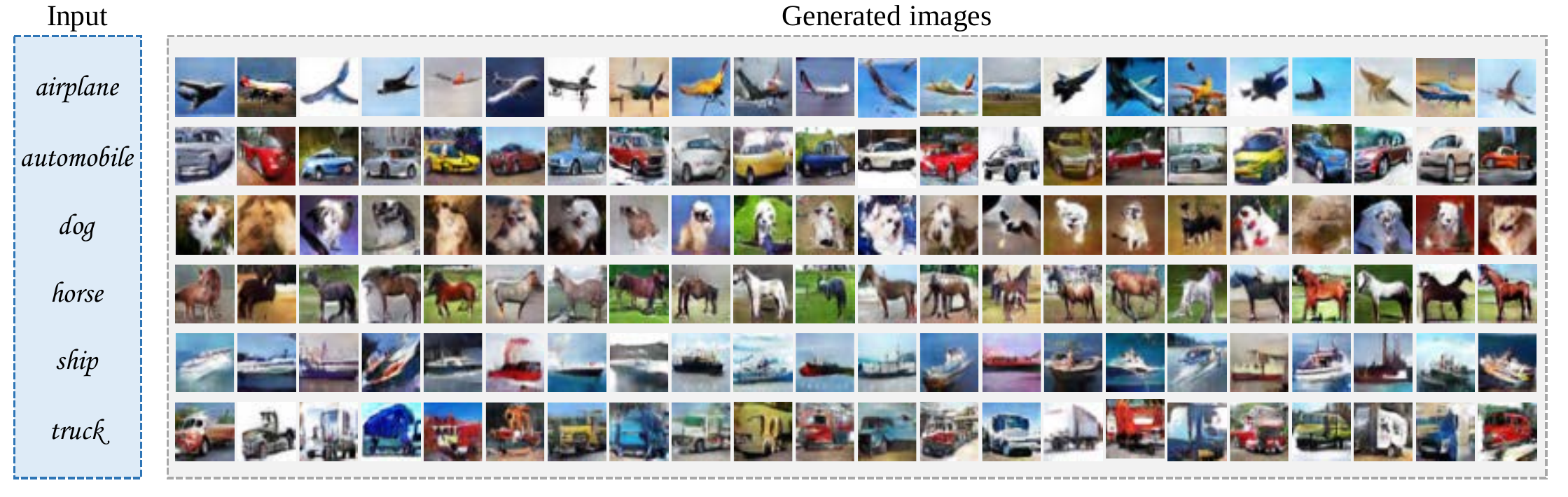}
\end{center}
\caption{\textbf{More categorical generation results of CIFAR-10.} We show the results of DCGAN~\cite{radford2015unsupervised} with the proposed mode seeking regularization term on categorical generation task.}
\label{fig:cifar}
\end{figure*}

\end{appendices}

\end{document}